\documentclass[10pt,twocolumn,letterpaper]{article}

\usepackage[pagenumbers]{iccv} 

\definecolor{iccvblue}{rgb}{0.21,0.49,0.74}

\usepackage[dvipsnames]{xcolor}
\usepackage{soul}
\usepackage[breakable,skins]{tcolorbox}
\usepackage{adjustbox}
\usepackage{lscape} 

\usepackage[T1]{fontenc}
\usepackage{newtxtext}
\usepackage[cmintegrals]{newtxmath}
\usepackage{bm}
\usepackage{amsmath}
\usepackage{amssymb}
\DeclareMathAlphabet{\mymathbb}{U}{BOONDOX-ds}{m}{n}

\usepackage{xspace}
\usepackage{multirow}
\usepackage{tabularx}
\usepackage{marvosym}
\usepackage{colortbl}
\usepackage{pifont}
\usepackage[normalem]{ulem}

\usepackage[accsupp]{axessibility}  

\usepackage{graphicx}
\graphicspath{{figures/}}

\usepackage{relsize}
\usepackage{nicematrix}

\usepackage{wrapfig}
\usepackage{makecell}

\usepackage{array}

\usepackage{url}

\usepackage{pifont}
\usepackage{tikz}
\usepackage{pgfplots}
\pgfplotsset{compat=1.18}

\usepackage{calc}

\newcommand{\lesspace}{\vspace{-0.3cm}}
\newcommand{\refapp}[1]{Supp.~\ref{sec:#1}}

\newcommand{\method}{RAXO\xspace}
\newcommand{\methodacro}{t\textbf{RA}ining-free adaptation for \textbf{X}-ray \textbf{O}pen-vocabulary detection\xspace}

\newcommand{\plusours}{ \ \ $\mathbf{+}$ \textbf{\method}}

\definecolor{goodgreen}{rgb}{0.0, 0.56, 0.0}
\definecolor{greengod}{RGB}{93, 184, 57}
\definecolor{badgray}{HTML}{666666}
\definecolor{badred}{RGB}{224, 53, 34}

\definecolor{tikzbaseline}{HTML}{FF2C71}
\definecolor{tikzours}{HTML}{027FFF}

\newcommand{\gooddelta}[1]{\scalebox{1.05}{$_{\textcolor{goodgreen}{\uparrow\bm{{#1}}}}$}}
\newcommand{\baddelta}[1]{\scalebox{1.0}{$_{\textcolor{badgray}{\downarrow{#1}}}$}}
\newcommand{\baddeltared}[1]{\scalebox{1.05}{$_{\textcolor{badred}{\downarrow\bm{{#1}}}}$}}

\newcommand{\gooddeltatext}[1]{\scalebox{0.90}{${\textcolor{goodgreen}{\uparrow}}$}${\textcolor{goodgreen}{{{#1}}}}$}
\newcommand{\baddeltaredtext}[1]{\scalebox{0.90}{${\textcolor{badred}{\downarrow}}$}${\textcolor{badred}{{{#1}}}}$}

\newcolumntype{R}{>{\raggedleft}p{0.043\textwidth}}
\newcolumntype{L}{>{\raggedright}p{0.043\textwidth}}
\newcolumntype{U}{>{\raggedleft}p{0.062\textwidth}}

\newcommand{\bI}{\mathbf{I}}
\newcommand{\bb}{\mathbf{b}}

\newcommand{\bz}{\mathbf{z}}

\newcommand{\bw}{\mathbf{w}}
\newcommand{\bW}{\mathbf{W}}

\newcommand{\classTrain}{\mathcal{C}^{\mathrm{train}}}

\newcommand{\classTest}{\mathcal{C}^{\mathrm{test}}}

\DeclareMathOperator*{\argmax}{arg\,max}

\usepackage[ruled,vlined,linesnumbered]{algorithm2e}

\newcommand{\bigO}[1]{\mathcal{O}(#1)}  

\usepackage[pagebackref,breaklinks,colorlinks,allcolors=iccvblue]{hyperref}

\def\confName{ICCV}
\def\confYear{2025}

\title{Superpowering Open-Vocabulary Object Detectors for X-ray Vision\texorpdfstring{\protect\vspace{-1.0em}}{}}

\author{
Pablo Garcia-Fernandez\textsuperscript{1}$^*$ \quad
Lorenzo Vaquero\textsuperscript{1,2} \quad
Mingxuan Liu\textsuperscript{3} \quad
Feng Xue\textsuperscript{3}$^*$  \\[-0.05em]
Daniel Cores\textsuperscript{1} \quad
Nicu Sebe\textsuperscript{3} \quad
Manuel Mucientes\textsuperscript{1} \quad
Elisa Ricci\textsuperscript{2,3} \\[0.4em]
\textsuperscript{1}University of Santiago de Compostela, Spain \quad \textsuperscript{2}Fondazione Bruno Kessler, Italy \\[-0.1em]
\textsuperscript{3}University of Trento, Italy \\[-0.05em]
{\tt\small \{pablogarcia.fernandez, daniel.cores, manuel.mucientes\}@usc.es \quad lvaquerootal@fbk.eu} \\[-0.25em]
{\tt\small \{mingxuan.liu, feng.xue, niculae.sebe, e.ricci\}@unitn.it}\texorpdfstring{{\small{\quad $^*$Corresponding authors}}}{}\texorpdfstring{\protect\vspace{-0.9em}}{}
}

\begin{document}
\maketitle
\begin{abstract}
Open-vocabulary object detection (OvOD) is set to revolutionize security screening by enabling systems to recognize any item in X-ray scans.
However, developing effective OvOD models for X-ray imaging presents unique challenges due to data scarcity and the modality gap that prevents direct adoption of RGB-based solutions.
To overcome these limitations, we propose \textbf{RAXO}, a training-free framework that repurposes off-the-shelf RGB OvOD detectors for robust X-ray detection.
RAXO builds high-quality X-ray class descriptors using a dual-source retrieval strategy.
It gathers relevant RGB images from the web and enriches them via a novel X-ray material transfer mechanism, eliminating the need for labeled databases.
These visual descriptors replace text-based classification in OvOD, leveraging intra-modal feature distances for robust detection.
Extensive experiments demonstrate that RAXO consistently improves OvOD performance, providing an average mAP increase of up to 17.0 points over base detectors.
To further support research in this emerging field, we also introduce DET-COMPASS, a new benchmark featuring bounding box annotations for over 300 object categories, enabling large-scale evaluation of OvOD in X-ray.
Code and dataset available at: \url{https://pagf188.github.io/RAXO/}.
\end{abstract}
\vspace{-0.3em}    
\section{Introduction}
\label{sec:intro}
Automated object detection technologies for X-ray imaging are essential to maintain public safety, enabling the identification of prohibited items at checkpoints in high-risk environments such as airports, train stations, museums and stadiums~\cite{Singh24}.
These systems improve security while simultaneously reducing the workload of human inspectors.

\begin{figure}[t]
\centering
\includegraphics[width=0.9\linewidth]{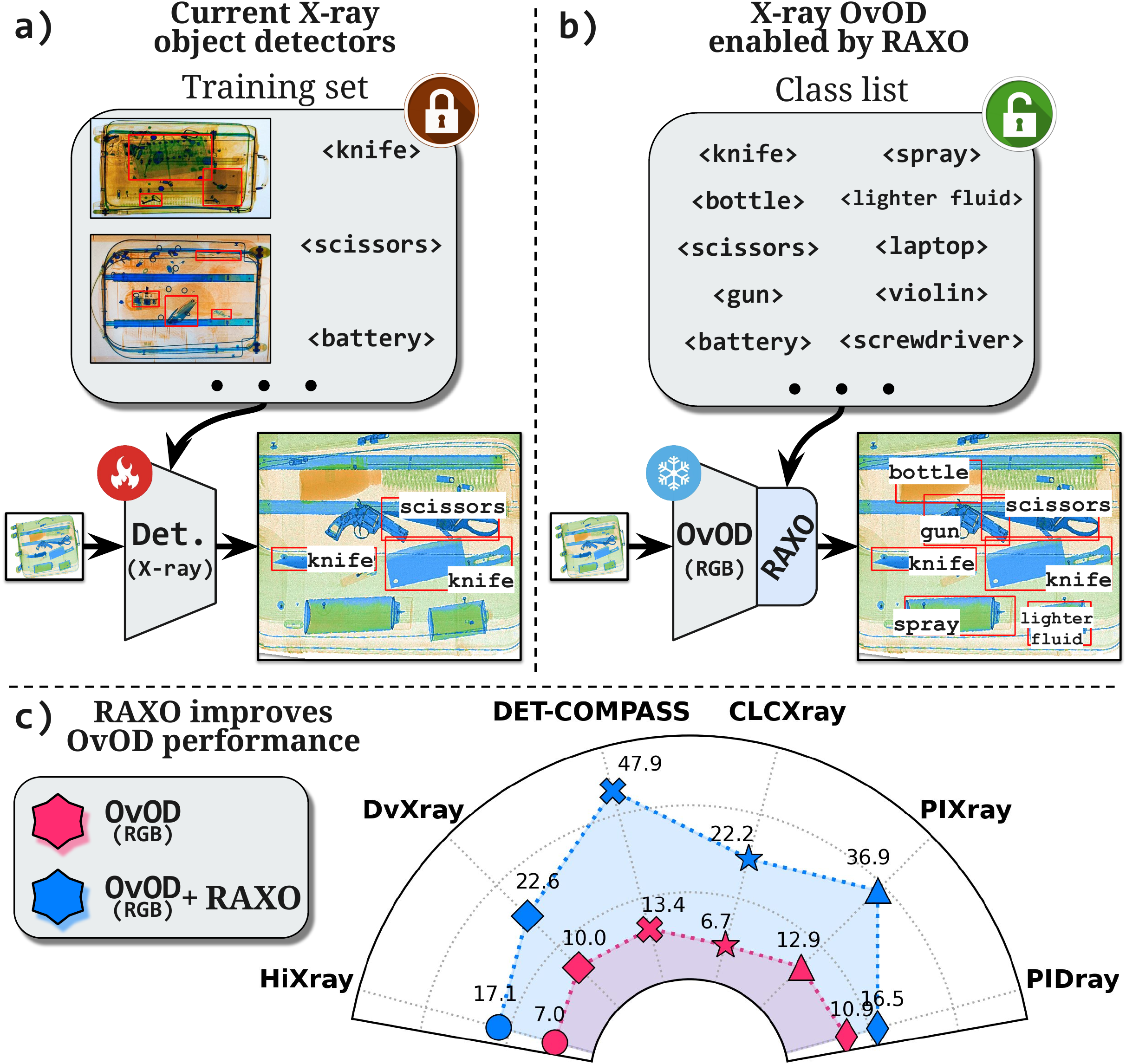}
\lesspace

\caption{
\textbf{(a)}~Traditional X-ray object detectors are constrained by the limited categories in their training datasets.
\textbf{(b)}~We introduce the task of open-vocabulary object detection (OvOD) for X-ray imaging and propose \method, a training-free method that adapts off-the-shelf RGB OvOD models to X-ray data.
\textbf{(c)}~\method greatly improves detection performance across multiple benchmarks.
}
\label{fig:teaser}
\vspace{-1.05em}
\end{figure}

Conventional X-ray object detectors rely on supervised learning~\cite{dvxray,chang2022detecting,10746383} and are inherently limited by the object categories present in their training datasets (see \cref{fig:teaser}\hyperref[fig:teaser]{a}).
This limitation is exacerbated by the high cost of X-ray machinery and the requirement for expert annotation, often restricting these systems to fewer than 20 object classes~\cite{xray_fsod}, thereby hindering broader real-world applications.

In light of the expanding diversity of man-made objects and evolving security demands, an open-vocabulary object detection (OvOD) framework capable of recognizing arbitrary X-ray object categories defined by the user is imperative.
Despite its importance, open-vocabulary detection in the X-ray domain remains largely unexplored.
Concurrent works aiming to extend beyond base categories have managed to generalize to at most four unseen classes~\cite{Lin24}, thereby highlighting the inherent challenges of the task.

Recent advances in OvOD for conventional RGB images have been driven by large-scale annotated datasets, which facilitate effective alignment between visual and textual features~\cite{detic,vldet,liu2025grounding}.
However, these advancements do not directly transfer to X-ray imagery.
In practice, applying state-of-the-art RGB-based OvOD models to X-ray scans leads to significant performance degradation, as illustrated in \cref{fig:teaser}\hyperref[fig:teaser]{c}.
Moreover, retraining these models on X-ray data is often impractical given the scarcity of large-scale annotated X-ray datasets.

Motivated by these challenges, this work opens up open-vocabulary object detection for X-ray imaging by repurposing robust, off-the-shelf RGB-based detectors.
To this end, we propose \textbf{\method} (\methodacro), a training-free method that seamlessly adapts RGB OvOD models to X-ray. (\cref{fig:teaser}\hyperref[fig:teaser]{b}).

We find that the main reason RGB-based OvOD detectors fail in the X-ray domain is the disruption of text-visual feature alignment. The appearance disparity between the same object in RGB and X-ray modalities causes textual embeddings, aligned with RGB features, to mismatch with X-ray visual features. \method adresses this issue on three simple steps: (1) \textit{visual sample acquisition}, which involves obtaining X-ray images that represent user-defined categories; (2) \textit{class descriptor modeling}, which uses these images to construct descriptors that effectively encode the visual appearance of each category within the X-ray domain; and (3) \textit{classifier construction}, where the computed descriptors are used to build a visual classifier that replaces the inter-modal (\ie, text-to-visual) classification of conventional OvODs. This new classifier exploits the fact that, despite the modality shift, intra-modal (\eg, visual-to-visual) feature distances remain reliable indicators for object identification. In this way \method enables any OvOD to successfully detect X-ray objects based on their inherent visual properties.

To facilitate comprehensive evaluation, we also introduce DET-COMPASS, a detection dataset comprising 370 distinct object classes with paired X-ray and RGB annotations. Extensive experiments demonstrate that our approach consistently improves detection performance by an average of \gooddeltatext{8.4} AP across multiple benchmarks and OvOD models, with gains that scale as the underlying detectors become more powerful (\cref{tab:main_exps}).
In summary, our contributions are:
\begin{itemize}
    \item We formally introduce the problem of open-vocabulary object detection for X-ray imagery without training, addressing a critical need for security in real-world scenarios.
    
    \item We introduce DET-COMPASS, a novel benchmark with bounding box annotations across 370 object categories, enabling standardized evaluation of OvOD methods in the X-ray modality.
    \item We propose \method, a novel training-free approach that repurposes off-the-shelf RGB-based OvOD models for X-ray detection by constructing robust visual descriptors. \method achieves new state-of-the-art results across multiple benchmarks.
\end{itemize}

\section{Related Work}
\label{sec:relatedwork}

\noindent\textbf{Open-vocabulary object detection} has advanced rapidly with the advent of Vision-Language Models (VLMs), enabling detectors to generalize beyond fixed categories.
OvOD detectors leverage weak supervision signals to improve detection accuracy. Based on the type of weak supervision utilized, OvOD methods can be categorized into (i) region-aware training, (ii) pseudo-labeling, (iii) knowledge distillation, and (iv) transfer learning.

Region-aware training methods aim to improve localization and feature representation by refining the alignment between image regions and their corresponding textual descriptions. Approaches such as DetCLIP \cite{yao2022detclip}, DetCLIPv2 \cite{yao2023detclipv2}, CORA \cite{wu2023cora}, and VLDet \cite{vldet} adopt this strategy.
Pseudo-labeling methods, rely on large pretrained VLMs to generate pseudo-labels, effectively expanding the training set.
Methods like RegionCLIP \cite{zhong2022regionclip}, PromptDet \cite{feng2022promptdet}, CoDET \cite{codet}, GLIP \cite{li2022grounded}, Detic \cite{detic}, and Grounding DINO \cite{liu2025grounding} follow this approach. Knowledge distillation techniques, such as BARON \cite{wu2023aligning}, DK-DETR \cite{li2023distilling}, CLIPSelf \cite{wuclipself}, and SIC-CADS \cite{fang2024simple}, employ VLMs as teachers in a teacher-student framework, transferring knowledge from the VLM image encoder to enhance the detector backbone. In contrast, transfer learning approaches integrate the VLM encoder directly, either through fine-tuning, as seen in OWL-ViT \cite{minderer2022simple}, or by freezing the encoder, as in F-VLM \cite{kuo2022f}.

Regardless of the training strategy, OvOD methods heavily depend on VLMs trained on large-scale datasets. However, such datasets are unavailable in the X-ray modality, making this approach impractical. To address this, \method seamlessly adapts existing RGB-based OvOD methods for X-ray object detection without requiring additional training.

\noindent\textbf{X-ray object detection.} Several datasets have been developed to tackle the challenge of object detection in X-ray imagery \cite{akcay2022towards,velayudhan2022recent}. SIXray \cite{miao2019sixray}, OPIXray \cite{OPIXray}, and CLCXray \cite{clcxray} focused on detecting occluded objects, introducing strategies to identify deliberately hidden prohibited object. To enhance dataset scale and diversity, PIDray \cite{wang2021towards} and HiXray \cite{HiXray} provided larger benchmarks, significantly increasing the number of annotated images. Meanwhile, PIXray \cite{pixray} pioneered the introduction of an X-ray segmentation dataset and proposed a real-time framework for segmenting prohibited objects, further advancing automated threat detection in security screening applications.

Previous X-ray object detectors \cite{eds,10082660,sima2024multi} considered mainly a closed-set paradigm,
thus being limited to localize objects of a small set of predefined categories.
More recently,
Lin \textit{et al.} \cite{Lin24} proposed fine-tuning a CLIP adapter to bridge the modality gap between CLIP’s training data and X-ray images, thereby achieving OvOD in X-ray data.
They demonstrated generalization on four novel object classes,
but both training and evaluation were limited by the lack of large-scale, well-annotated X-ray data.
To address these issues,
\method leverages data retrieval from the web to expand detection capabilities to a broader range of common objects.
Furthermore, our proposed re-labeled DET-COMPASS dataset enables evaluation across a wider variety of classes.

\section{DET-COMPASS}
\label{sec:dataset}

\begin{table}[!t]
\centering
\resizebox{1.0\columnwidth}{!}{%
\begin{tabular}{@{}l@{\hskip 8pt}c@{\hskip 6pt}cc@{\hskip 6pt}c@{}}
\toprule
        & \textbf{Venue} & \textbf{Images} & \textbf{Classes} & \textbf{Modality}\\
\hline
DvXray~\cite{dvxray}  & TIFS'24 & 32,000 & 15 & X-ray \\
PIXray~\cite{pixray}  & TMM'22 & 5,046 & 15 & X-ray \\
CLCXray~\cite{clcxray}  & TIFS'22 & 9,565 & 12 & X-ray \\
FSOD~\cite{xray_fsod}  & ACMMM'22 & 12,333 & 20 & X-ray \\
EDS~\cite{eds}  & CVPR'22 & 14,219 & 10 & X-ray \\
PIDray~\cite{wang2021towards}  & ICCV'21 & 47,677 & 12 & X-ray \\
HiXray~\cite{HiXray}  & ICCV'21 & 45,365 & 8 & X-ray \\
\midrule
\textbf{DET-COMPASS (Ours)} &  -- & \textbf{1,928} & \textbf{370} & \textbf{X-ray+RGB} \\
\bottomrule
\end{tabular}
}
\lesspace
\caption{\textbf{Existing X-ray detection datasets}.
DET-COMPASS is the dataset with the highest number of categories and the only one providing pixel-level alignment between X-ray and RGB data.}
\lesspace
\vspace*{-4pt}
\label{tab:datasets}
\end{table}

Object detection in security X-ray scans has advanced significantly in recent years. However, evaluating OvOD detectors in this modality remains challenging due to the limited number of annotated object categories in existing X-ray benchmarks. For instance, the largest X-ray detection dataset, FSOD~\cite{xray_fsod}, includes annotations for only 20 classes (see \cref{tab:datasets}). This limitation severely constrains the comprehensive evaluation of OvOD methods, which require a broad and diverse category set (or say vocabulary) to assess generalization to unseen object semantics.

To address this gap, we introduce \textit{DET-COMPASS}, a novel benchmark that repurposes the COMPASS-XP classification dataset~\cite{compass} for object detection through meticulous \emph{manual bounding box annotation}. DET-COMPASS comprises 370 distinct object classes, offering \emph{an order-of-magnitude increase in vocabulary size} over previous X-ray detection benchmarks. Additionally, it provides pixel-aligned RGB images, ensuring precise spatial correspondence across modalities and facilitating the development of multimodal models. Each object is also labeled with a visibility attribute, indicating whether it produces a discernible signature in the X-ray spectrum. Further details are provided in \cref{sec:det_compass_details}.

As summarized in \cref{tab:datasets}, DET-COMPASS sets a new standard in class diversity and uniquely integrates multimodal, pixel-aligned X-ray and RGB data. The dataset will be released under an open license, serving as a valuable resource for advancing OvOD research in security screening and industrial inspection.

\begin{figure*}[t]
\centering
\includegraphics[width=0.95\linewidth]{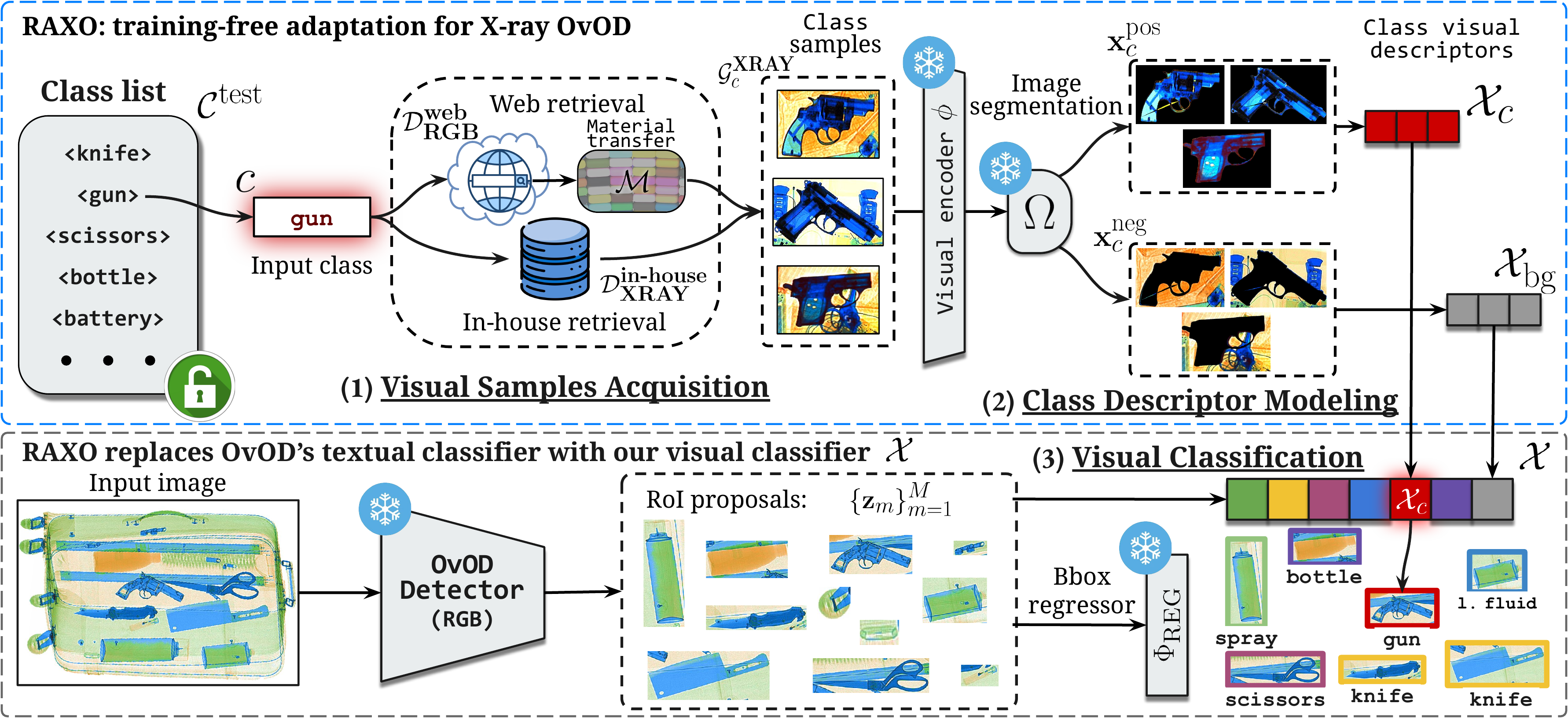}
\lesspace
\caption{
\textbf{Architecture of \method}.
For a given user-defined class $c \in \classTest$, \method first retrieves its corresponding X-ray images $\mathcal{G}_{c}^{XRAY}$ from in-house and web sources, using its \textbf{(1)} \emph{Visual Samples Acquisition} pipeline (\cref{sec:vea}).
Following this, \method extracts the features of the images and segments them with its \textbf{(2)} \emph{Class Descriptor Modeling} module (\cref{sec:vcm}), creating ensemble visual descriptors for the class $\mathcal{X}_c$ and the background $\mathcal{X}_\text{bg}$.
Finally, the text-based classifier from the baseline RGB OvOD detector is replaced with our \textbf{(3)} \emph{Visual-based Classifier} (\cref{sec:rec}) $\mathcal{X}$, which yields accurate predictions on the X-ray modality.
}
\label{fig:main_architecture}
\end{figure*}
\section{OvOD for X-ray Imaging} 
\label{sec:problemdefinition}

\paragraph{Preliminaries: OvOD in the RGB domain.}
Most RGB OvOD detectors~\cite{detic, zhu2024survey} follow a two-stage pipeline.
During training, a region proposal network (RPN) is learned to yield a set of $M$ proposals by $\{\bz_m\}_{m=1}^{M} = \Phi_\textsc{RPN}(\bI^{RGB})$, where $\bz_m \in \mathbb{R}^D$ is a $D$-dimensional region-of-interest (RoI) feature embedding.
Then,
a bounding box regressor predicts coordinates for each proposed region via $\hat{\bb}_m = \Phi_{\textsc{REG}}(\bz_m)$.
A set of text-based classifiers $\bW\!=\!\{\bw_{c}|\bw_{c}\!\in\!\mathbb{R}^D\}_{c=1}^{|\classTrain|}$ are used to compute classification scores for each region as $\langle \bw_{c}, \bz_m \rangle$,
where $\langle \cdot, \cdot \rangle$ is the cosine similarity function
and $\classTrain$ denotes the training vocabulary.
{In this way,
each region's class is determined by the class with the highest score.
Here, the classifier $\bW$ is constructed by encoding class names in $\classTrain$ using a pre-trained VLM text encoder,
e.g., CLIP~\cite{radford2021learning}.
During training, OvOD models update all parameters while keeping $\bW$ frozen.}
This enables RGB-region-class alignment by leveraging large-scale RGB data and the pre-aligned vision-language semantic space of VLMs,
facilitating open-vocabulary inference with any test vocabulary $\classTest$. The vocabularies $\classTrain$ and $\classTest$ may be disjoint or overlapping.

\vspace{3pt}
\noindent\textbf{Problem formulation.}
{In this work, we study open-vocabulary object detection (OvOD) in X-ray modality.}
Specifically, given an input X-ray image $\bI$ and a vocabulary $\classTest$ defined by users at test time,
{an X-ray OvOD detector $\mathcal{F}_{\text{X-ray}}$ aims to detect objects specified in $\classTest$ from $\bI$}
(\eg, a ``Power bank'' in a passenger's backpack scan at an airport security checkpoint).
{Theoretically, this detection process can be formulated as $\mathcal{F}_{\text{X-ray}}\!\!:\! \bI \rightarrow \{(\bb_m, c_m)\}_{m=1}^{M}$,
where $\bb_m \in \mathbb{R}^4$ denotes the coordinates of each bounding box, and $c_m \in \classTest$ denotes the class label of each bounding box.}

X-ray OvOD presents significant challenges due to the following reasons:
{\textit{i)} Directly applying a pre-trained RGB OvOD detectors to X-ray images leads to suboptimal performance due to the visual modality gap, as shown in \cref{fig:teaser};
\textit{ii)} The scarcity of large-scale security X-ray datasets limits the application of RGB OvOD training techniques.
These techniques typically rely on extensive image-text annotated data for strong supervision.}
{To tackle these challenges,
we introduce RAXO,
a plug-and-play module that adapts any off-the-shelf RGB OvOD detectors to X-ray modalities.
RAXO requires no training or in-domain detection annotations.
Next, we present our approach.}

\section{\method: Training-free Modality Adaptation}
\label{sec:methodDescription}

As illustrated in \cref{fig:main_architecture}, \method enables X-ray OvOD by constructing high-quality visual descriptors, $\mathcal{X}_c$, for each class $c$ in the user-defined vocabulary, $\classTest$. To achieve this, \method first \textbf{acquires} X-ray samples in an open-ended manner, leveraging both \textit{in-house} and \textit{web-based} retrieval sources (\cref{sec:vea}). Then, it extracts the features of these samples and segments the relevant information, \textbf{modeling} the visual descriptor $\mathcal{X}_c$ (\cref{sec:vcm}). Once the visual descriptors are constructed offline, they can be directly applied to any \textit{off-the-shelf} OvOD detector by replacing the conventional text-based classifier $\mathbf{W}$, with our \textbf{visual-based classifier} $\mathcal{X}$ (\cref{sec:rec}). Thus, \method effectively overcomes the misalignment between X-ray features and text semantics, enabling OvOD detectors pre-trained with RGB data to accurately identify X-ray objects based on their intrinsic visual characteristics.

\subsection{Visual Sample Acquisition}
\label{sec:vea}
Obtaining informative and representative images is crucial for generating robust visual descriptors. To this end, we propose a \textit{dual-source} acquisition pipeline that retrieves a gallery of relevant X-ray samples $\mathcal{G}_{c}^{\text{XRAY}}$ for each user-specified class $c \in \mathcal{C}^{\text{test}}$. This pipeline consists of two modules: an \textit{in-house} retrieval module and a \textit{web-powered} retrieval module.

\vspace{3pt}
\noindent\textbf{In-house retrieval.}  
Given an in-domain X-ray dataset, $\mathcal{D}^{\text{in-house}}_{\text{XRAY}}$, containing categories $\mathcal{C}^{\text{in-house}}$, our sample acquisition pipeline first attempts to retrieve the $K$ most relevant images for the user-specified class $c$ based on class name matching: $\mathcal{G}_{c}^{\text{XRAY}} \!\!=\!\! \left\{ \mathbf{I} | (\mathbf{I}, c) \!\in\! \mathcal{D}^{\text{in-house}}_{\text{XRAY}}\right\}^{K}$.

However, since the user-defined vocabulary $\classTest$ is open-ended, we cannot assume that every user-specified category will be present in $\mathcal{C}^{\text{in-house}}$.
Consequently, for some categories, the retrieved set $\mathcal{G}_{c}^{\text{XRAY}}$ may be empty. To overcome this limitation and fully support open-vocabulary user input, \method further incorporates a novel \textit{web-powered} retrieval module. This module retrieves RGB images from the web and applies a material transfer mechanism to synthesize X-ray-style representations, which we introduce next.

\vspace{3pt}
\noindent\textbf{Web-powered retrieval.}  
To obtain high-quality visual samples for a category $c$ that is not present in the in-house vocabulary $\mathcal{C}^{\text{in-house}}$, we leverage the vast availability of web-based RGB image data $\mathcal{D}^{\text{web}}_{\text{RGB}}$ as an open-ended auxiliary source. Specifically, we perform text-based web retrieval using the class name $c$ as a search query, retrieving the top $K$ results from Google Images as $\widetilde{\mathcal{G}}_{c}^{\text{web}} = \left\{ \mathbf{I^{\text{RGB}}} \mid (\mathbf{I^{\text{RGB}}}, c) \in \mathcal{D}^{\text{web}}_{\text{RGB}}\right\}^{K}$.

The raw web-retrieved results $\widetilde{\mathcal{G}}_{c}^{\text{web}}$ are often noisy and may not always contain clear instances of the target class $c$. To refine these results, we apply a filtering step using an RGB OvOD detector $\mathcal{F}_{\text{RGB}}$. Specifically, we discard images in the raw web-retrieved results where class $c$ is \textit{not} confidently detected, retaining only those where the detection confidence exceeds a threshold $\tau$ as $\mathcal{G}_{c}^{\text{web}} = \operatorname{Filter}(\widetilde{\mathcal{G}}_{c}^{\text{web}}, \mathcal{F}_{\text{RGB}}, c, \tau)$.

\begin{figure}[t]
\centering
\includegraphics[width=0.99\linewidth]{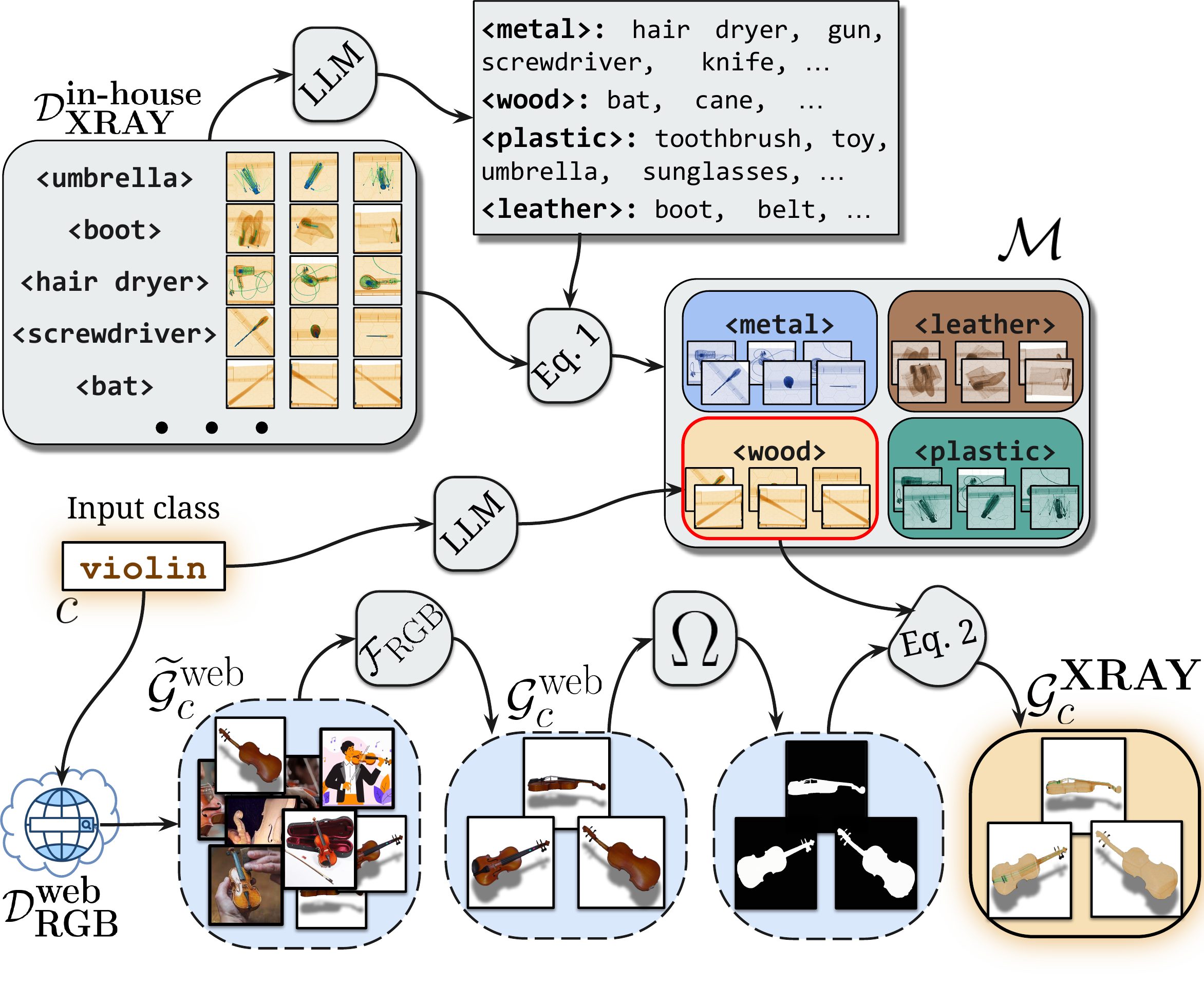}
\lesspace
\caption{
\textbf{Web-powered retrieval and material-transfer mechanism} for the class ``\textit{violin}''. We retrieve violin samples from the web, filter them using $\mathcal{F}_{\text{RGB}}$, and inpaint the retrieved appearance into the object masks to generate synthetic X-ray samples.
}
\label{fig:visualExamplarAcq}
\end{figure}

\vspace{3pt}
\noindent\textbf{Material-transfer mechanism for web-retrieved images.}
The substantial visual disparity between RGB and X-ray modalities prevents direct use of web-retrieved RGB samples $\mathcal{G}_{c}^{\text{web}}$ for constructing X-ray class descriptors. Style transfer methods~\cite{styleshot} fall short in bridging this gap, as they fail to capture the underlying material properties of objects (as shown in \cref{tab:domains_ablation}). To address this, we introduce a novel material-transfer mechanism for generating synthetic X-ray samples from web-retrieved images. As shown in \cref{fig:visualExamplarAcq}, our approach consists of two key steps: \textit{i)} constructing an offline material database $\mathcal{M}$ that encapsulates the expected X-ray appearance of various \textit{materials}, and \textit{ii)} adapting RGB samples $\mathcal{G}_{c}^{\text{web}}$ to X-ray style by applying the corresponding material properties from $\mathcal{M}$.

To construct \(\mathcal{M}\), we employ a Large Language Model (LLM) to cluster the class names of \(\mathcal{C}^{\text{in-house}}\) into subsets $\mathcal{C}_m \subset \mathcal{C}^{\text{in-house}}$, where $\mathcal{C}_m$ contains classes that share the same $m$ material  -- \ie, ``$\mathcal{C}_\textit{metal} = \{ \text{\textit{gun}}, \text{\textit{knife}}, \text{\textit{fork}} \}$'' or ``$\mathcal{C}_\textit{leather} = \{ \text{\textit{boot}}, \text{\textit{belt}} \}$''. Each material $m$ is then associated with the corresponding set of images $\mathcal{K}_m = \{\, \mathbf{I} \mid (\mathbf{I}, c_k) \in \mathcal{D}^{\text{in-house}}_{\text{XRAY}}, \, c_k \in C_m\}$. These images are used to compute the appearance, $\mathcal{A}_m\in \mathbb{R}^3$, of the material as:
\begin{equation}
    \mathcal{A}_m = \frac{1}{|\mathcal{K}_m|} \sum_{\mathbf{I} \in \mathcal{K}_m} \left( \frac{\sum_{i,j}^{H,W}( \mathbf{I}_{i,j} \odot \Omega(\mathbf{I}_{i,j})) }{\sum \Omega(\mathbf{I})} \right),
\end{equation}
where $\odot$ denotes element-wise multiplication and \(\Omega: \mathbf{I} \to \{ 0, 1 \}^{H \times W}\) is a segmentation function that produces a binary mask, to isolate the object of interest in $\mathbf{I}$.
The resulting material-appearance pairs $(m, \mathcal{A}_m)$ collectively form our material database $\mathcal{M}$.

Once \(\mathcal{M}\) is computed offline, it can be used to adapt the web-retrieved RGB samples, $\mathcal{G}_{c}^{\text{web}}$, to the X-ray modality. To achieve this, we first prompt the LLM to link the class \(c\) with its corresponding material \(m_c \in \mathcal{M}\) (\eg, $c=\text{``violin''}$ is linked with the material $m_{c}=\text{``wood''}$).
Subsequently, the X-ray appearance $\mathcal{A}_m^c$ of the linked material is used to render the characteristic X-ray style on each web-retrieved RGB samples as:
\begin{equation}
\mathcal{G}^{\text{XRAY}}_{c} = \left\{\Omega(\mathbf{I^{\text{RGB}}}) \odot (\mathcal{A}_m^c \cdot \mathbf{1}) \mid \mathbf{I^{\text{RGB}}} \in \mathcal{G}_{c}^{\text{web}} \right\},
\end{equation}

Finally, the X-ray visual gallery for every class freely specified by the user is built as $\mathcal{G} = \bigcup_{c \in \classTest} \mathcal{G}^{\text{XRAY}}_{c}$ through our dual-source retrieval pipeline, seamlessly integrating high-quality in-house X-ray samples with synthetic images derived from web data.

\subsection{Class Descriptor Modeling}
\label{sec:vcm}

The effectiveness of \method hinges on leveraging its visual gallery $\mathcal{G}$ to construct a robust visual descriptor $\mathcal{X}_c$ that accurately captures the visual X-ray properties of class $c$. A naïve approach, such as averaging the feature representations of all $\mathcal{G}_c^{\text{XRAY}}$ instances, fails to account for intra-class variability (\eg, distinct shapes of utility knives versus chef knives) and does not differentiate between foreground and background regions.
To address these limitations, we propose a novel class descriptor modeling strategy shown in \cref{fig:main_architecture}\hyperref[fig:main_architecture]{(2)}.

First, we process each sample \(\mathbf{I} \in \mathcal{G}_{c}^{XRAY}\) with \(\Omega\) to separate the X-ray object from its background.
Simultaneously, a feature extractor \(\phi\) extracts per-patch embeddings \(\phi(\mathbf{I}) \in \mathbb{R}^{H' \times W' \times D}\), where \(H' \times W'\) represents the number of spatial tokens, and \(D\) is the embedding dimension. Following~\cite{karazija2024diffusion}, we resize (denoted by $\zeta$) the segmentation mask to match the spatial dimensions of the feature map, and subsequently we use it to compute both a positive and a negative prototype.
The \emph{positive prototype}, designed to capture the significant visual features of the object without background interference, is computed as the average embedding over the foreground region as:
\begin{equation}
\label{eq:pos_prot}
\mathbf{x}^{\text{pos}}_\mathbf{I} = \frac{\sum_{i,j}^{H',W'} \zeta(\Omega(\mathbf{I}_{i,j})) \odot \phi(\mathbf{I}_{i,j})}{\sum \zeta(\Omega(\mathbf{I}))},
\end{equation}
On the other hand, the \emph{negative prototype} is obtained as the average embedding over the complementary region (background) as:
\begin{equation}
\label{eq:neg_prot}
\mathbf{x}^{\text{neg}}_\mathbf{I} = \frac{\sum_{i,j}^{H',W'} (1-\zeta(\Omega(\mathbf{I_{i,j}}))) \odot \phi(\mathbf{I_{i,j}})}{\sum (1-\zeta(\Omega(\mathbf{I})))}.
\end{equation}
Subsequently, to construct the final visual descriptor $\mathcal{X}_c$ of class \(c\), we compute the average positive prototype $\widebar{\mathbf{x}}^{\text{pos}}_c$ from $\mathcal{G}_{c}^{XRAY}$ and unite it with all the individual positive prototypes as:
\begin{equation}
\mathcal{X}_c = [\widebar{\mathbf{x}}^{\text{pos}}_c, \left\{ \mathbf{x}^{\text{pos}}_\mathbf{I} \mid \mathbf{I} \in \mathcal{G}_c \right\}].
\end{equation}
This formulation effectively captures both fine-grained object details and a holistic class-level representation, handling intra-class variability.
Additionally, we also construct a global \emph{background descriptor} by using the negative prototypes across the entire visual sample gallery as $\mathcal{X}_{bg} = [\widebar{\mathbf{x}}^{\text{neg}},  \left\{ \mathbf{x}^{\text{neg}}_\mathbf{I} \mid \mathbf{I} \in \mathcal{G} \right\}]$.
This negative prototype is used to further improve detection reliability by filtering out low-quality proposals (\eg, a region proposed on the background) during inference.
A key advantage of this approach is its modularity:
visual descriptors $\mathcal{X}_c$ are computed \textit{offline} and can be incrementally expanded with new object categories, seamlessly integrating with the OvOD paradigm without training or requiring  X-ray detection data.

\subsection{Classification is All You Need}
\label{sec:rec}

As shown in \cref{fig:main_architecture}\hyperref[fig:main_architecture]{(3)}, once the visual descriptors are constructed for each class in the user-specified vocabulary, along with the background class, the \method classifier $\mathcal{X} = \{ {[\mathcal{X}_c, \mathcal{X}_{bg}] \mid c \in \classTest}\}$ can be directly applied to any OvOD detector to classify proposals $\mathbf{z_m}$ as:

\begin{equation}
    \hat{c} = \argmax_{c \in \mathcal{C}^{\text{test}^\prime}} \; \max_{\mathbf{x} \in \mathcal{X}} \langle \mathbf{z}_m, \mathbf{x} \rangle,
\end{equation}
where $\mathcal{C}^{\text{test}^\prime}$ denotes the test-time vocabulary $\classTest$ extended with the additional ``background'' class. If a proposal matches the background, it is removed.
This simple-yet-effective strategy enables any pre-trained OvOD detectors to achieve strong performance on X-ray object detection with minimal modifications and no need for re-training.

\vspace{3pt}
\noindent\textbf{Descriptor consistency criterion.} To further reduce incorrect proposals from the OvOD detector, \method introduces a novel \textit{Descriptor Consistency Criterion} (DCC) to enforce class consistency in predictions. DCC evaluates how well a proposal aligns with its predicted class relative to others, suppressing weakly aligned proposals. For a proposal $\mathbf{z}_m$, we measure its similarity to the closest prototype as $s_1 = \max_{\mathbf{x} \in \mathcal{X}_{\hat{c}}} \langle \mathbf{z}_m, \mathbf{x} \rangle$. We also compute its mean similarity to the average prototypes of all other classes as $s_2 = \operatorname{avg}_{c \in \mathcal{C}^{\text{test}}, c \ne \hat{c}} \langle \mathbf{z}_m, \bar{\mathbf{x}}^{\text{pos}}_c \rangle$. The difference $\Delta = s_1 - s_2$ serves as a confidence measure, where a higher value indicates greater alignment with its predicted class than with any other. Proposals with $\Delta$ below a threshold $\sigma$ are discarded.

\section{Experiments}
\label{sec:experiments}

\paragraph{Evaluation protocol.}
Conventional OvOD models are typically assessed using a \textit{Cross-Dataset Transfer Evaluation} (CDTE) protocol, where the model is trained on one dataset and evaluated on a different dataset in a zero-shot manner~\cite{zhu2024survey}.
In contrast, our setting involves a more challenging scenario where both the datasets and their underlying modalities differ.
To address this, we introduce a novel \textit{Cross-Modality Transfer Evaluation} (\textbf{CMTE}) protocol.
Under CMTE, an OvOD model is trained on a \emph{source RGB} dataset and subsequently evaluated on \emph{target X-ray} datasets without any additional training or fine-tuning.
For performance assessment, we employ the standard MS COCO~\cite{coco} metrics: AP, AP50, and AP75.

\vspace{3pt}
\noindent\textbf{Implementation details.}
Unless stated otherwise, \method's default configuration employs SAM~2~\cite{sam2} as the object segmentation module $\Omega$, ViT-B/14~\cite{Dosovitskiy21} pretrained with DINOv2 \cite{dinov2} as the visual encoder $\phi$, and GPT-4~\cite{OpenAI2023} as the LLM.
We set the consistency threshold to $\sigma = 0.15$, the similarity threshold to $\tau=0.5$, and use $K=30$ samples to construct the visual descriptors. $\mathcal{D}^{in-house}_{\text{XRAY}}$ is composed of hold-out samples sourced from the evaluation X-ray datasets and is completely disjoint from the test split.
$\mathcal{D}^{web}_{\text{RGB}}$ is constructed by collecting images from the web using the public Google Custom Search API~\cite{googleapi}.
A detailed analysis of the impact of hyperparameter choices is provided in~\cref{sec:ablation}. Further implementation details can be found in \refapp{details2}.

\vspace{3pt}
\noindent\textbf{Datasets and baselines.}
We evaluate \method by integrating it with four different state-of-the-art OvOD detectors that were exclusively trained on RGB images: GroundingDINO~\cite{liu2025grounding}, Detic~\cite{detic}, VLDet~\cite{vldet}, and CoDet~\cite{codet}.
Our CMTE evaluation is performed across six diverse X-ray datasets: PIXray~\cite{pixray}, PIDray~\cite{wang2021towards}, CLCXray~\cite{clcxray}, DvXray~\cite{dvxray}, HiXray~\cite{HiXray}, and our proposed DET-COMPASS, which together comprise 343 visible unique classes and over 140k images. A detailed description of these datasets is provided in \cref{tab:datasets}.

\subsection{Open-Vocabulary Detection Results}
\label{sec:ovod_results}
\begin{table*}[!t]
\setlength\extrarowheight{-3pt}
\centering
\resizebox{0.93\linewidth}{!}{%
\begin{tabular}{c@{\hskip 4pt}c@{\hskip 6pt}l|U@{\hskip 2pt}LR@{\hskip 2pt}LR@{\hskip 2pt}LR@{\hskip 2pt}LR@{\hskip 2pt}LR@{\hskip 2pt}L|r@{\hskip 2pt}l}
\toprule

\multicolumn{2}{c}{$\bm{\mathcal{G}}$} & \multicolumn{1}{c|}{\textbf{Method}} & \multicolumn{2}{c}{\textbf{D-COMPASS}} & \multicolumn{2}{c}{\textbf{PIXray}} & \multicolumn{2}{c}{\textbf{PIDray}} & \multicolumn{2}{c}{\textbf{CLCXray}} & \multicolumn{2}{c}{\textbf{DvXray}} & \multicolumn{2}{c|}{\textbf{HiXray}} & \multicolumn{2}{c}{\textbf{Avg.}} \\

\midrule
\midrule

 & & G-DINO~\cite{liu2025grounding} & \multicolumn{2}{c}{13.4} & \multicolumn{2}{c}{12.9} & \multicolumn{2}{c}{10.9} & \multicolumn{2}{c}{6.7} & \multicolumn{2}{c}{10.0} & \multicolumn{2}{c|}{7.0} & \multicolumn{2}{c}{10.2} \\

\midrule

\multirow{5}{*}{
    \makecell{
    \rotatebox[origin=c]{45}{$\mathcal{D}^{\text{in-h}}_{\text{{\smaller{XRAY\hspace{-50pt}}}}}$}\\[-7pt]
    \tikz[baseline=(current bounding box.center)]{
      \draw[->, thick] (0, 0.6) -- (0, 0.1); 
    }\\[-2pt]
    \rotatebox[origin=c]{45}{$\mathcal{D}^{\text{web}}_{\text{{\smaller{RGB}}}}$}
    }%
  }%
& \textit{100/0} &  & \textbf{47.9} & \gooddelta{34.5} & \textbf{36.9} & \gooddelta{24.0} & \textbf{16.5} & \gooddelta{5.6} & \textbf{22.2} & \gooddelta{15.5} & \textbf{22.6} & \gooddelta{12.6} & \textbf{17.1} & \gooddelta{10.1} & \textbf{27.2} & \gooddelta{17.0} \\

& \textit{80/20} &  & 41.0 & \gooddelta{27.6} & 33.8 & \gooddelta{20.9} & 15.4 & \gooddelta{4.5} & 18.0 & \gooddelta{11.3} & 21.0 & \gooddelta{11.0} & 14.5 & \gooddelta{7.5} & 24.0 & \gooddelta{13.8} \\

& \textit{50/50} & \plusours & 31.4 & \gooddelta{18.0} & 25.4 & \gooddelta{12.5} & 15.5 & \gooddelta{4.6} & 17.0 & \gooddelta{10.3} & 16.1 & \gooddelta{6.1} & 13.4 & \gooddelta{6.4} & 19.8 & \gooddelta{9.6} \\

& \textit{20/80} &  & 20.5 & \gooddelta{7.1} & 21.6 & \gooddelta{8.7} & 13.9 & \gooddelta{3.0} & 10.0 & \gooddelta{3.3} & 15.0 & \gooddelta{5.0} & 9.8 & \gooddelta{2.8} & 15.1 & \gooddelta{4.9} \\

& \textit{0/100} &  & 14.0 & \gooddelta{0.6} & 16.1 & \gooddelta{3.2} & 13.4 & \gooddelta{2.5} & 7.1 & \gooddelta{0.4} & 12.4 & \gooddelta{2.4} & 7.9 & \gooddelta{0.9} & 11.8 & \gooddelta{1.6} \\

\midrule
\midrule

 & & VLDet~\cite{vldet} & \multicolumn{2}{c}{10.6} & \multicolumn{2}{c}{9.8} & \multicolumn{2}{c}{6.9} & \multicolumn{2}{c}{4.4} & \multicolumn{2}{c}{7.4} & \multicolumn{2}{c|}{5.1} & \multicolumn{2}{c}{7.4} \\

\midrule

\multirow{5}{*}{
    \makecell{
    \rotatebox[origin=c]{45}{$\mathcal{D}^{\text{in-h}}_{\text{{\smaller{XRAY\hspace{-50pt}}}}}$}\\[-7pt]
    \tikz[baseline=(current bounding box.center)]{
      \draw[->, thick] (0, 0.6) -- (0, 0.1); 
    }\\[-2pt]
    \rotatebox[origin=c]{45}{$\mathcal{D}^{\text{web}}_{\text{{\smaller{RGB}}}}$}
    }%
  }%
& \textit{100/0} &  & \textbf{36.4} & \gooddelta{25.8} & \textbf{32.3} & \gooddelta{22.5} & \textbf{11.7} & \gooddelta{4.8} & \textbf{15.4} & \gooddelta{11.0} & \textbf{20.1} & \gooddelta{12.7} & \textbf{14.8} & \gooddelta{9.7} & \textbf{21.8} & \gooddelta{14.4} \\

& \textit{80/20} &  & 31.8 & \gooddelta{21.2} & 29.2 & \gooddelta{19.4} & 11.0 & \gooddelta{4.1} & 12.7 & \gooddelta{8.3} & 16.8 & \gooddelta{9.4} & 13.1 & \gooddelta{8.0} & 19.1 & \gooddelta{11.7} \\

& \textit{50/50} & \plusours & 23.7 & \gooddelta{13.1} & 24.0 & \gooddelta{14.2} & 10.4 & \gooddelta{3.5} & 11.1 & \gooddelta{6.7} & 12.1 & \gooddelta{4.7} & 11.2 & \gooddelta{6.1} & 15.4 & \gooddelta{8.0} \\

& \textit{20/80} &  & 16.2 & \gooddelta{5.6} & 21.6 & \gooddelta{11.8} & 9.4 & \gooddelta{2.5} & 5.2 & \gooddelta{0.8} & 10.6 & \gooddelta{3.2} & 9.3 & \gooddelta{4.2} & 12.1 & \gooddelta{4.7} \\

& \textit{0/100} &  & 11.1 & \gooddelta{0.5} & 14.1 & \gooddelta{4.3} & 8.9 & \gooddelta{2.0} & 4.4 & \gooddelta{0.0} & 9.0 & \gooddelta{1.6} & 8.3 & \gooddelta{3.2} & 9.3 & \gooddelta{1.9} \\

\midrule
\midrule

 & & Detic~\cite{detic} & \multicolumn{2}{c}{11.5} & \multicolumn{2}{c}{9.3} & \multicolumn{2}{c}{7.1} & \multicolumn{2}{c}{4.7} & \multicolumn{2}{c}{7.0} & \multicolumn{2}{c|}{4.8} & \multicolumn{2}{c}{7.4} \\

\midrule

\multirow{5}{*}{
    \makecell{
    \rotatebox[origin=c]{45}{$\mathcal{D}^{\text{in-h}}_{\text{{\smaller{XRAY\hspace{-50pt}}}}}$}\\[-7pt]
    \tikz[baseline=(current bounding box.center)]{
      \draw[->, thick] (0, 0.6) -- (0, 0.1); 
    }\\[-2pt]
    \rotatebox[origin=c]{45}{$\mathcal{D}^{\text{web}}_{\text{{\smaller{RGB}}}}$}
    }%
  }%
& \textit{100/0} &  & \textbf{35.3} & \gooddelta{23.8} & \textbf{27.3} & \gooddelta{18.0} & \textbf{11.3} & \gooddelta{4.2} & \textbf{14.0} & \gooddelta{9.3} & \textbf{19.4} & \gooddelta{12.4} & \textbf{14.2} & \gooddelta{9.4} & \textbf{20.3} & \gooddelta{12.9} \\

& \textit{80/20} &  & 30.7 & \gooddelta{19.2} & 23.9 & \gooddelta{14.6} & 10.8 & \gooddelta{3.7} & 12.3 & \gooddelta{7.6} & 18.0 & \gooddelta{11.0} & 12.1 & \gooddelta{7.3} & 18.0 & \gooddelta{10.6} \\

& \textit{50/50} & \plusours & 24.4 & \gooddelta{12.9} & 19.5 & \gooddelta{10.2} & 10.3 & \gooddelta{3.2} & 9.2 & \gooddelta{4.5} & 14.6 & \gooddelta{7.6} & 11.0 & \gooddelta{6.2} & 14.8 & \gooddelta{7.4} \\

& \textit{20/80} &  & 16.4 & \gooddelta{4.9} & 15.2 & \gooddelta{5.9} & 9.6 & \gooddelta{2.5} & 8.0 & \gooddelta{3.3} & 12.7 & \gooddelta{5.7} & 9.9 & \gooddelta{5.1} & 12.0 & \gooddelta{4.6} \\

& \textit{0/100} &  & 11.9 & \gooddelta{0.4} & 13.4 & \gooddelta{4.1} & 9.1 & \gooddelta{2.0} & 5.2 & \gooddelta{0.5} & 9.4 & \gooddelta{2.4} & 7.9 & \gooddelta{3.1} & 9.5 & \gooddelta{2.1} \\

\midrule
\midrule

 & & CoDet~\cite{codet} & \multicolumn{2}{c}{8.4} & \multicolumn{2}{c}{7.3} & \multicolumn{2}{c}{5.7} & \multicolumn{2}{c}{3.1} & \multicolumn{2}{c}{5.6} & \multicolumn{2}{c|}{3.4} & \multicolumn{2}{c}{5.6} \\

\midrule

\multirow{5}{*}{
    \makecell{
    \rotatebox[origin=c]{45}{$\mathcal{D}^{\text{in-h}}_{\text{{\smaller{XRAY\hspace{-50pt}}}}}$}\\[-7pt]
    \tikz[baseline=(current bounding box.center)]{
      \draw[->, thick] (0, 0.6) -- (0, 0.1); 
    }\\[-2pt]
    \rotatebox[origin=c]{45}{$\mathcal{D}^{\text{web}}_{\text{{\smaller{RGB}}}}$}
    }%
  }%
& \textit{100/0} &  & \textbf{35.8} & \gooddelta{27.4} & \textbf{27.9} & \gooddelta{20.6} & \textbf{10.3} & \gooddelta{4.6} & \textbf{14.8} & \gooddelta{11.7} & \textbf{17.6} & \gooddelta{12.0} & \textbf{13.2} & \gooddelta{9.8} & \textbf{19.9} & \gooddelta{14.3} \\

& \textit{80/20} &  & 32.2 & \gooddelta{23.8} & 25.1 & \gooddelta{17.8} & 9.5 & \gooddelta{3.8} & 12.0 & \gooddelta{8.9} & 15.4 & \gooddelta{9.8} & 11.7 & \gooddelta{8.3} & 17.7 & \gooddelta{12.1} \\

& \textit{50/50} & \plusours & 24.0 & \gooddelta{15.6} & 20.0 & \gooddelta{12.7} & 9.5 & \gooddelta{3.8} & 9.2 & \gooddelta{6.1} & 11.5 & \gooddelta{5.9} & 9.9 & \gooddelta{6.5} & 14.0 & \gooddelta{8.4} \\

& \textit{20/80} &  & 17.8 & \gooddelta{9.4} & 14.8 & \gooddelta{7.5} & 8.5 & \gooddelta{2.8} & 5.1 & \gooddelta{2.0} & 9.4 & \gooddelta{3.8} & 8.1 & \gooddelta{4.7} & 10.6 & \gooddelta{5.0} \\

& \textit{0/100} &  & 12.2 & \gooddelta{3.8} & 11.5 & \gooddelta{4.2} & 8.1 & \gooddelta{2.4} & 4.0 & \gooddelta{0.9} & 6.9 & \gooddelta{1.3} & 6.5 & \gooddelta{3.1} & 8.2 & \gooddelta{2.6} \\

\bottomrule
\end{tabular}
}
\vspace{-0.2cm}
\caption{
\textbf{X-ray OvOD performance under the Cross-Modality Transfer Evaluation (CMTE) setting} on DET-COMPASS (ours), PIXray~\cite{pixray}, PIDray~\cite{wang2021towards}, CLCXray~\cite{clcxray}, DvXray~\cite{dvxray}, and HiXray~\cite{HiXray} datasets.
We integrate \method into different baselines using different gallery $\mathcal{G}$ compositions, from using only $\mathcal{D}^{\text{in-house}}_{\text{XRAY}}$ data (100/0) to exclusively $\mathcal{D}^{web}_{\text{RGB}}$ samples (0/100).
\method consistently improves the performance of all baseline OvOD detectors across every dataset. AP is used. 
}
\vspace{-0.3cm}
\label{tab:main_exps}
\end{table*}

\noindent\textbf{\method equips OvOD models with X-ray vision.}
We evaluate the X-ray adaptation capabilities of \method on the aforementioned datasets and baselines in \cref{tab:main_exps}.
Since \method is training-free, we report the results directly on the test splits, accounting for all available categories.
Furthermore, we analyze the influence of the gallery composition by varying $\mathcal{G}$ from a \textit{0/100} configuration (\ie, all samples in $\mathcal{G}$ are retrieved from the web, as $\mathcal{C}^{\text{test}} \cap \mathcal{C}^{\text{in-house}} = \emptyset$) to a \textit{100/0} configuration (\ie, all samples in $\mathcal{G}$ are retrieved from the in-house database, as $\mathcal{C}^{\text{test}} \subset \mathcal{C}^{\text{in-house}}$).
Each experiment is repeated three times with different random distributions of in-domain and web categories, and we report the averaged results.

As shown in \cref{tab:main_exps}, off-the-shelf OvOD detectors trained on RGB images perform poorly on X-ray data. In contrast, even without access to any in-domain data from $\mathcal{C}^{\text{test}}$ (\textit{0/100} setting), \method enhances baseline OvOD methods, yielding an average improvement of \gooddeltatext{2.1} points.
This gain is further amplified when in-domain samples are incorporated into $\mathcal{G}$, which facilitates the construction of more robust materials in $\mathcal{M}$ and yields higher-quality visual descriptors.
Notably, even a limited number of in-domain samples (\textit{20/80} setting) results in an average improvement of \gooddeltatext{4.8} points.
Moreover, increasing the in-domain samples (\textit{50/50} setting) yields even more significant gains, particularly in datasets with a large number of classes, such as DET-COMPASS, where \method offers an average boost of \gooddeltatext{14.9} points.
Finally, when $\mathcal{G}$ consists entirely of in-domain samples, \method delivers an average improvement of \gooddeltatext{14.7} points across all OvOD methods and datasets, all in a training-free manner.

\vspace{3pt}
\noindent\textbf{\method works with any OvOD detector.}
The consistent gains of \method across all evaluated detectors in \cref{tab:main_exps}, along with its simple integration (\cref{sec:rec}), confirm the generalization of our approach to any RGB OvOD model.
Moreover, \method scales with the baseline method, yielding larger gains when integrated with more robust detectors (\eg, \method enhances G-DINO \gooddeltatext{9.4} points on average across all $\mathcal{G}$ settings, compared to \gooddeltatext{7.5} points for Detic).
This highlights the advantage of our training-free method, which can leverage the rapid advancements in RGB OvOD detectors without needing large amounts of labeled X-ray data.

\begin{table}[!t]
\setlength\extrarowheight{-2pt}
\centering
\resizebox{0.97\columnwidth}{!}{%
\begin{tabular}{l|llllllllllllllllllllllll|c}
\toprule

& \multicolumn{25}{c}{\textbf{PIXray}} \\

& \multicolumn{6}{c}{Scissors} & \multicolumn{6}{c}{Wrench} & \multicolumn{6}{c}{Battery} & \multicolumn{6}{c|}{Pliers} & \textbf{AP50} \\
\midrule
\midrule

OVXD$^{\dagger}$ & \multicolumn{6}{c}{16.9} & \multicolumn{6}{c}{46.2} & \multicolumn{6}{c}{14.6} & \multicolumn{6}{c|}{6.4} & 21.0 \\
\midrule

BARON + \textbf{\method} & \multicolumn{6}{c}{18.2} & \multicolumn{6}{c}{39.4} & \multicolumn{6}{c}{8.3} & \multicolumn{6}{c|}{22.0} & 22.0 \\

CoDet + \textbf{\method} & \multicolumn{6}{c}{43.6} & \multicolumn{6}{c}{52.6} & \multicolumn{6}{c}{9.0} & \multicolumn{6}{c|}{18.6} & 30.9 \\

Detic + \textbf{\method} & \multicolumn{6}{c}{46.3} & \multicolumn{6}{c}{60.5} & \multicolumn{6}{c}{7.9} & \multicolumn{6}{c|}{36.6} & 37.8 \\

VLDet + \textbf{\method} & \multicolumn{6}{c}{48.3} & \multicolumn{6}{c}{62.7} & \multicolumn{6}{c}{10.6} & \multicolumn{6}{c|}{38.5} & 40.0 \\

G-DINO + \textbf{\method} & \multicolumn{6}{c}{49.7} & \multicolumn{6}{c}{61.6} & \multicolumn{6}{c}{11.3} & \multicolumn{6}{c|}{51.5} & \textbf{43.5} \\

\bottomrule
\end{tabular}
}
\vspace{-0.15cm}
\caption{
\textbf{Comparison under OVXD~\cite{Lin24} setting}, where the methods do not have access to in-domain data from the displayed categories.
$^{\dagger}$OVXD is a supervised method explicitly trained on X-ray images for alignment.
AP50 is used.
}
\vspace{-0.45cm}
\label{tab:supervised_comparison}
\end{table}

\vspace{3pt}
\noindent\textbf{\method surpasses training-based approaches.}
We also compare \method with OVXD~\cite{Lin24}, the only concurrent work that extends X-ray detection beyond base categories.
OVXD is a \textit{fully-supervised} method trained directly on X-ray data, and its generalization capabilities are evaluated on a hold-out set comprising only four categories (\ie, \texttt{Scissors}, \texttt{Wrench}, \texttt{Battery}, and \texttt{Pliers}).
In \cref{tab:supervised_comparison}, we follow this setting and evaluate \method on PIXray~\cite{pixray}, excluding the aforementioned classes from the in-house dataset $\mathcal{D}^{\text{in-house}}_{\text{XRAY}}$.
Our results demonstrate that \method outperforms OVXD even when both approaches utilize the same BARON detector~\cite{wu2023aligning}.
Notably, unlike OVXD, our method does \underline{not} retrain BARON or any other component. Yet, our training-free \method performs remarkably well, achieving an AP50 of 43.5 when paired with G-DINO~\cite{liu2025grounding}.

\subsection{Ablation Study}
\label{sec:ablation}
In this section we study the core components of \method.
We conduct our experiments on PIXray~\cite{pixray}, using GroundingDINO~\cite{li2022grounded} as our baseline OvOD. Consistent findings are reported in \refapp{results2} and \ref{sec:qualitative_method}.

\begin{table}[!t]
\setlength\extrarowheight{-2pt}
\centering
\resizebox{0.95\columnwidth}{!}{%
\begin{tabular}{l|c@{\hskip 3pt}c@{\hskip 3pt}c@{\hskip 2pt}|c@{\hskip 4pt}c@{\hskip 4pt}c@{\hskip 4pt}}
\toprule

& \textbf{Mod.} & \textbf{Filt.} & \textbf{Trans.} & \multicolumn{1}{c}{\textbf{AP}} & \multicolumn{1}{c}{\textbf{AP50}} & \multicolumn{1}{c}{\textbf{AP75}} \\

\midrule
\midrule

G-DINO \cite{liu2025grounding} & & & & 12.9 & 14.9 & 13.4\\

\midrule

\multirow{5}{*}{
    \plusours
  }%
& X-ray & & & 7.9\baddeltared{5.0} & 9.8\baddeltared{5.1} & 8.3\baddeltared{5.1}\\

& RGB & & &13.9\gooddelta{1.0} & 17.4\gooddelta{2.5} & 14.3\gooddelta{0.9}\\

& RGB & \checkmark & & 14.3\gooddelta{1.4} & 17.6\gooddelta{2.7} & 14.7\gooddelta{1.3}\\

& RGB & \checkmark & $\mathcal{S}$ & 14.6\gooddelta{1.7} & 18.3\gooddelta{3.4} & 15.1\gooddelta{1.7}  \\	

& RGB & \checkmark & $\mathcal{M}$ & \textbf{16.1}\gooddelta{3.2} & \textbf{19.8}\gooddelta{4.9} & \textbf{16.8}\gooddelta{3.4} \\	

\bottomrule
\end{tabular}
}
\vspace{-0.15cm}
\caption{
\textbf{Ablation of web-based retrieval} on PIXray~\cite{pixray} using only web-based samples (setting \textit{0/100}).
\textbf{Mod.} indicates retrieval of a specific modality from the web, \textbf{Filt.} refers to filtering retrieved images with $\mathcal{F}_{\text{RGB}}$, and \textbf{Trans.} specifies style-transfer via: $\mathcal{S}$ a diffusion-based method~\cite{styleshot} or $\mathcal{M}$ our material-transfer mechanism.
}
\vspace{-0.4cm}
\label{tab:domains_ablation}
\end{table}

\vspace{3pt}
\noindent\textbf{Web-based samples.}
\cref{tab:domains_ablation} presents an ablation study on the components of our web-based visual sample acquisition pipeline (\cref{sec:vea}).
Notably, directly retrieving X-ray samples from the web ($\text{Mod.} = \text{X-ray}$) leads to a \baddeltaredtext{5.0} point decrease in AP. In contrast, retrieving RGB images from the web ($\text{Mod.} = \text{RGB}$) provides a modest AP improvement of \gooddeltatext{1.0} point. We hypothesize that directly using X-ray examples obtained from the web is ineffective due to the limited availability of high-quality X-ray images online. Filtering the retrieved RGB images using $\mathcal{F}_{\text{RGB}}$ increases AP by an additional \gooddeltatext{0.4} points.
Nonetheless, they still exhibit a significant visual gap compared to their X-ray counterparts.
A style-transfer approach using StyleShot~\cite{styleshot} ($\text{Trans.} = \mathcal{S}$) provides a marginal AP gain of \gooddeltatext{0.3} points, likely due to its limited capacity to understand the intrinsic material properties of the objects.
Conversely, our proposed material-transfer mechanism ($\text{Trans.} = \mathcal{M}$) yields a substantial AP improvement of \gooddeltatext{1.8} points, underscoring its effectiveness in bridging the X-ray modality gap.

\begin{table}[!t]
\setlength\extrarowheight{-2pt}
\centering
\resizebox{0.95\columnwidth}{!}{%
\begin{tabular}{ll@{\hskip 4pt}|c@{\hskip 4pt}c@{\hskip 4pt}c@{\hskip 4pt}c@{\hskip 2pt}|c@{\hskip 4pt}c@{\hskip 4pt}c@{\hskip 4pt}}
\toprule

& & $\bm{\Omega}$ & $\bm{\mathbf{x}^{\text{neg}}}$ & $\bm{\mathcal{X}}$ & \textbf{DCC} & \textbf{AP} & \textbf{AP50} & \textbf{AP75} \\

\midrule
\midrule

\multicolumn{2}{l|}{G-DINO~\cite{liu2025grounding}} & & & & & 12.9 & 14.9 & 13.4\\

\midrule

\multirow{6}{*}{
    \plusours
  }%
& & & & & & 25.2\gooddelta{12.3} & 31.4\gooddelta{16.5} & 26.2\gooddelta{12.8}\\
& & \checkmark & & & & 27.1\gooddelta{14.2} & 33.6\gooddelta{18.7} & 28.4\gooddelta{15.0} \\
& & \checkmark & \checkmark & & & 27.8\gooddelta{14.9} & 34.5\gooddelta{19.6} & 29.3\gooddelta{15.9} \\

& & \checkmark & & \checkmark & & 27.5\gooddelta{14.6} & 34.0\gooddelta{19.1} & 28.8\gooddelta{15.4} \\

& & \checkmark & \checkmark & \checkmark & & 28.5\gooddelta{15.6} & 35.3\gooddelta{20.4} & 29.9\gooddelta{16.5} \\
& & \checkmark & \checkmark & \checkmark & \checkmark & \textbf{36.9}\gooddelta{24.0} & \textbf{45.0}\gooddelta{30.1} & \textbf{39.0}\gooddelta{25.6}\\

\bottomrule
\end{tabular}
}
\vspace{-0.2cm}
\caption{
\textbf{Ablation of class representation construction} on PIXray~\cite{pixray} using in-domain samples (setting \textit{100/0}). $\bm{\Omega}$ indicates building positive prototypes using segmentation masks.
$\bm{\mathbf{x}^{\text{neg}}}$ indicates use of negative prototypes.
$\bm{\mathcal{X}}$ indicates ensemble multiple prototypes in class descriptors.
$DCC$ indicates use of prototype consistency criterion.
}
\vspace{-0.4cm}
\label{tab:sam_ablation}
\end{table}

\vspace{3pt}
\noindent\textbf{Class descriptor modeling.}
\cref{tab:sam_ablation} presents an ablation study analyzing the impact of the various components used to construct class representations.
As shown, using a simple per-class average of all the features from each sample already results in a significant AP improvement of \gooddeltatext{12.3} points over the baseline OvOD.
Refining the representation by applying an object segmentation method to isolate the foreground features ($\Omega$) further boosts the AP by \gooddeltatext{1.9} points.
Moreover, leveraging background features to construct negative prototypes ($\mathbf{x}^{\text{neg}}$) adds an additional \gooddeltatext{0.7} points.

Replacing the averaged class prototypes with our proposed visual descriptor ($\mathcal{X}$) yields another \gooddeltatext{0.7} point improvement, demonstrating its enhanced capability to capture intra-class visual variability.
Finally, incorporating our descriptor consistency criterion to suppress low-quality proposals ($DCC$) provides a further AP increase of \gooddeltatext{8.4} points, underscoring the effectiveness of our proposed framework.

\begin{figure}[t]
\centering
\includegraphics[width=0.98\linewidth]{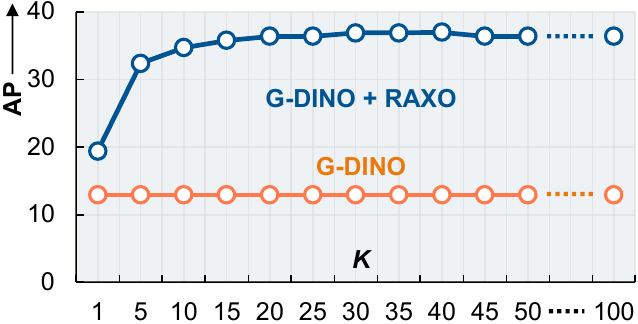}
\vspace{-0.2cm}
\caption{
\textbf{Impact of $\bm{K}$ in class representations} evaluated on PIXray~\cite{pixray} using a G-DINO~\cite{liu2025grounding} baseline in the \textit{100/0} setting.
}
\vspace{-0.1cm}
\label{fig:ablation_k}
\end{figure}

\begin{figure}[t]
\centering
\includegraphics[width=0.98\linewidth]{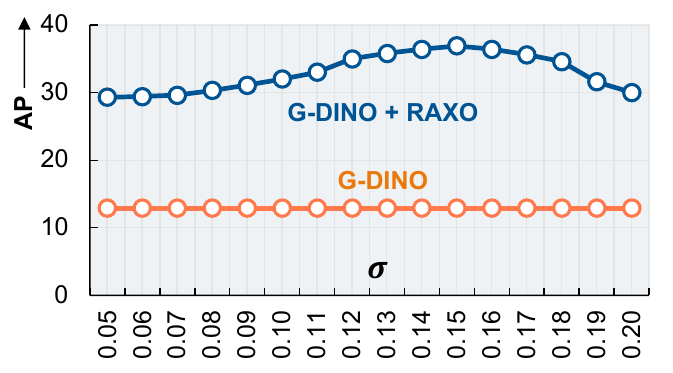}
\vspace{-0.2cm}
\caption{
\textbf{Impact of $\bm{\sigma}$ in Descriptor Consistency Module} on PIXray~\cite{pixray} using a G-DINO~\cite{liu2025grounding} baseline in the \textit{100/0} setting.}
\vspace{-0.3cm}
\label{fig:ablation_ths_uncer}
\end{figure}

\vspace{3pt}
\noindent\textbf{Hyperparameter study.}
\cref{fig:ablation_k} analyzes the effect of varying the number of samples $\bm{K}$ used to construct the visual descriptors.
Even with just a single sample per class, \method improves AP by \gooddeltatext{6.5} points over the baseline.
As $K$ increases, performance continues to improve, reaching saturation at $K=30$ with a total AP gain of \gooddeltatext{24.0} points.

Furthermore, \cref{fig:ablation_ths_uncer} explores the influence of the threshold $\bm{\sigma}$ in the \textit{Descriptor Consistency Module}.
Higher values of $\sigma$ impose stricter consistency requirements, demanding greater separation between the predicted class and other classes before accepting a proposal.
We observe that $\sigma = 0.15$ provides an optimal balance between false positives and false negatives, yielding an AP of $36.9$ points.

\section{Conclusions}
\label{sec:conclusions}

In this work, we pioneered the task of open-vocabulary object detection (OvOD) for X-ray imagery, a challenge shaped by the unique characteristics of X-ray data and the scarcity of annotated examples.
To address this, we introduced \method, a training-free method that repurposes off-the-shelf RGB-based OvOD detectors for the X-ray domain.
By leveraging intra-modal feature distances, a novel material-transfer mechanism, and robust class descriptor modeling, \method effectively bridges the modality gap between RGB and X-ray imagery.
Our extensive experiments, conducted across multiple benchmarks and the newly proposed DET-COMPASS dataset, demonstrate that \method consistently enhances detection performance, achieving average mAP improvements of up to $17.0$ points over baseline methods.

Looking ahead, the modularity of \method opens promising avenues for future research, including further refinement of visual descriptors and adaptation to additional modalities.
Overall, our work not only sets a new state-of-the-art for X-ray open-vocabulary detection but also lays the groundwork for a thriving research direction in this emerging domain.

\section*{Acknowledgements}
We thank CINECA and the ISCRA initiative for the availability of high-performance computing resources.
This work was partially supported by the EU HORIZON IAMI (HORIZON-CL3-2023-FCT-01-04-101168272) project, the EU HORIZON ELIAS (HORIZON-CL4-2022-HUMAN-02-101120237) project, the EU HORIZON ELLIOT (HORIZON-CL4-2024-HUMAN-03-101214398) project, the MUR PNRR FAIR (PE00000013) project funded by the NextGenerationEU, the Spanish Ministerio de Ciencia e Innovación (grant numbers PID2020-112623GB-I00, PID2023-149549NB-I00), and the Galician Consellería de Cultura, Educación e Ordenación Universitaria (2024-2027 ED431G-2023/04).
Some of these grants are co-funded by the European Regional Development Fund (ERDF).
Pablo Garcia-Fernandez is supported by the Spanish Ministerio de Universidades under the FPU national plan (grant number FPU21/05581).
{
    \small
    \bibliographystyle{ieeenat_fullname}
    \bibliography{main}
}

\clearpage
\setcounter{page}{1}
\maketitlesupplementary
\appendix

This supplementary material is organized into the following sections: \refapp{ethics} outlines ethical considerations related to our work; \refapp{repro} provides the reproducibility statement; \refapp{det_compass_details} describes the main characteristics and construction process of our proposed dataset, DET-COMPASS; \refapp{details2} presents additional technical implementation details of \method; \refapp{results2} and \refapp{sup_perclasaap} offer further analyses of RAXO's effectiveness; and \refapp{sup_qual_mt} and \refapp{qualitative_method} present insights into its performance through qualitative examples.

\section{Ethics Statement}
\label{sec:ethics}
We do not anticipate any immediate negative societal impact from our work. However, we encourage future researchers building upon this study to exercise the same level of caution we have maintained, recognizing that \method\ has the potential to be applied for both beneficial and harmful purposes.

The primary motivation behind our research is to enhance open-world perception in X-ray prohibited object detection, addressing the growing diversity of objects in security screening. By improving detection capabilities, our work aims to strengthen public safety in critical security scenarios. Notably, the proposed pipeline and model can be executed entirely on local systems, ensuring that user or institutional privacy remains well protected.

For evaluation, we rely on publicly available, well-established benchmarks, strictly adhering to their licensing terms. Regarding the new DET-COMPASS benchmark introduced in this work, we source images from the publicly available COMPASS-XP~\cite{compass} X-ray classification dataset, complying fully with its license. Our contribution lies in providing additional bounding box annotations to COMPASS-XP through our human annotation efforts. Importantly, we do not introduce or collect any new images. The human annotation process for DET-COMPASS was conducted following the approval of our institution’s ethics board after a thorough committee review.

Lastly, for web-retrieved images, we only retain those explicitly permitted for non-commercial use in this project. Each retrieved image was manually reviewed, ensuring that none contain private information such as human faces or vehicle license plates. We will release our proposed benchmark and prototypes under an appropriate license.

\section{Reproducibility Statement}
\label{sec:repro}
Upon publication, we will make all necessary resources available to facilitate the full reproduction of our experimental results. This includes the source code, precise prompts, and benchmark datasets with their splits. Our proposed framework, \method, is developed using \textit{open-source}, \textit{publicly accessible} models and data, reinforcing its reproducibility. A comprehensive breakdown of our pipeline’s construction is provided in \cref{sec:methodDescription}. Additionally, our supplementary material offers further implementation specifics, including the exact prompts, to assist practitioners in replicating our approach effortlessly.  By offering detailed methodological explanations, extensive experimental results, and a fully open-source framework and data, we aim to ensure that our work is easily reproducible, empowering researchers and practitioners to adapt our method across diverse applications.

\begin{figure}[t]
\centering
\includegraphics[width=\linewidth]{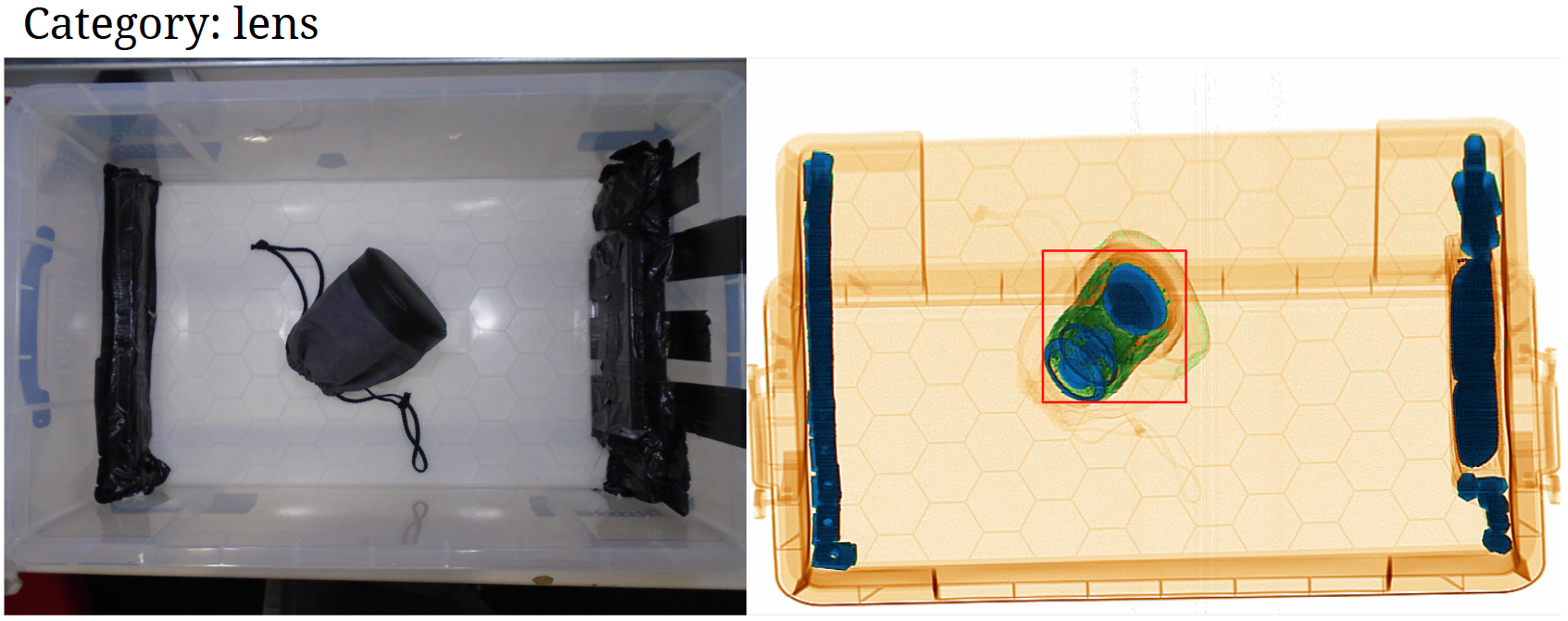}
\caption{\textbf{Occluded RGB object.} In this pair of images, the object \texttt{lens} is completely occluded in the RGB image, preventing the annotation of a bounding box
}
\label{fig:compass_qualitative_p1}
\end{figure}

\begin{figure}[t]
\centering
\includegraphics[width=\linewidth]{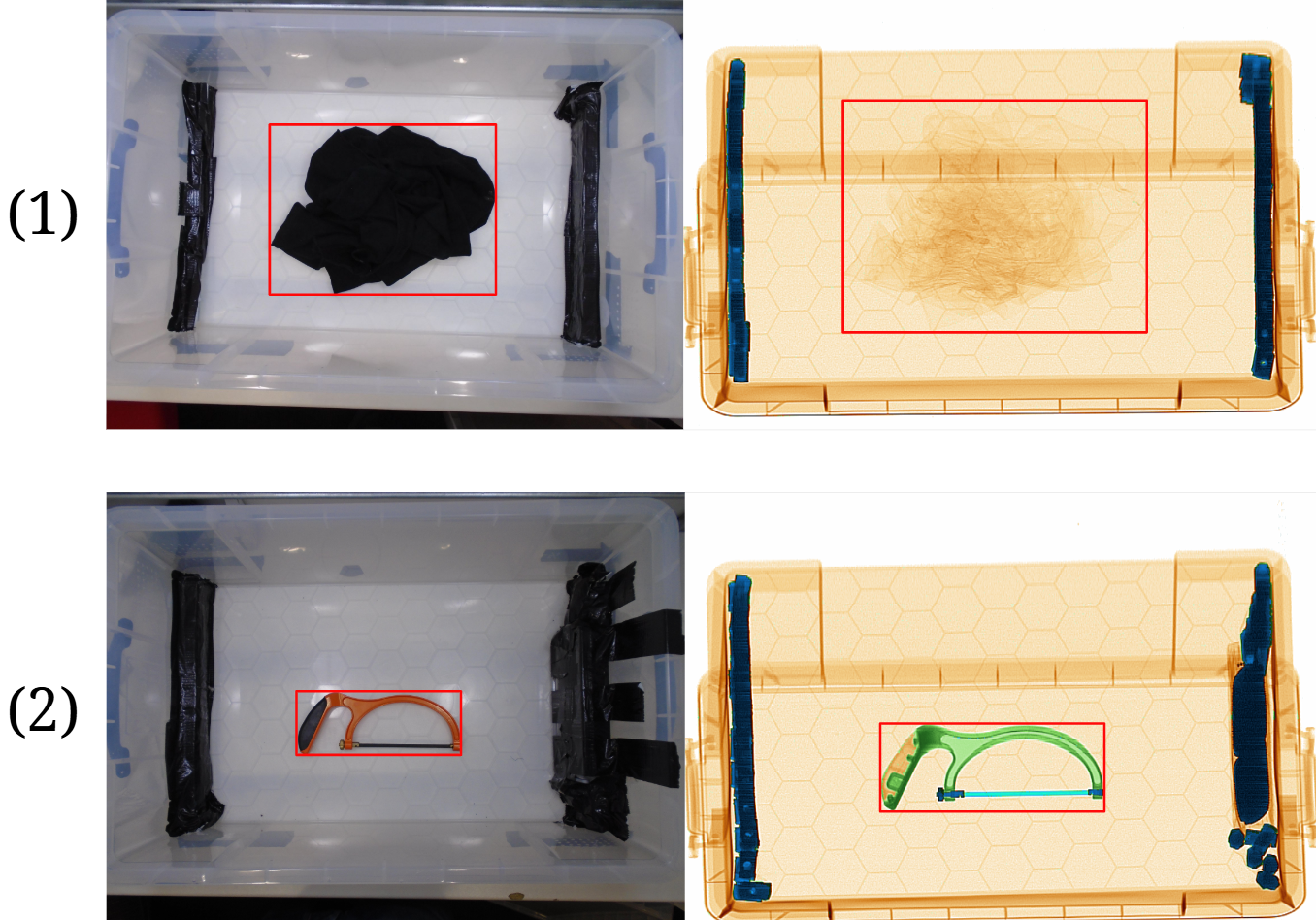}
\caption{\textbf{Visibility attribute.} In (1), the cardigan does not have a discernible signature in the X-ray spectrum, thus \texttt{visible=False}. In (2), the hacksaw does, so \texttt{visible=True}.
}
\label{fig:compass_qualitative_p2}
\end{figure}

\section{DET-COMPASS Details}
\label{sec:det_compass_details}

\begin{figure*}[t]
\centering
\includegraphics[width=0.89\linewidth]{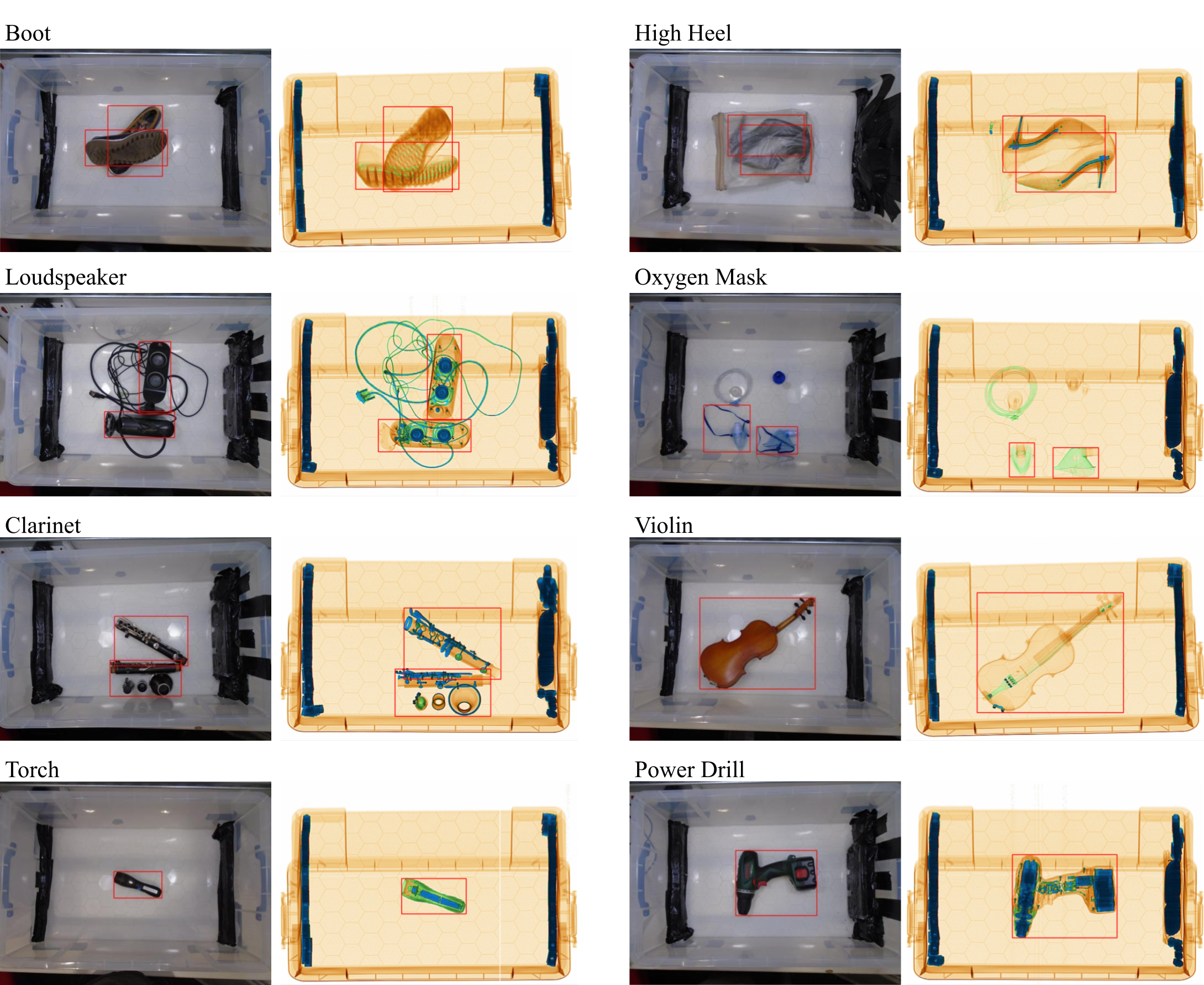}
\caption{Examples from our \textbf{DET-COMPASS dataset}, showing RGB-X-ray pairs with annotated bounding boxes.
}
\label{fig:compass_qualitative}
\end{figure*}

To construct our new DET-COMPASS dataset, we sourced images from the publicly available COMPASS-XP~\cite{compass} dataset. Both the images and their metadata are licensed under the Creative Commons Attribution 4.0 International License, permitting unrestricted use for research and commercial applications. COMPASS-XP comprises 1,928 image pairs, each consisting of an X-ray image captured with a Gilardoni FEP ME 536 scanner and a corresponding natural image taken with a Sony DSC-W800 digital camera. A key limitation of COMPASS-XP is that it provides only classification labels and the (RGB X-ray) pairs are not spatially aligned.

Our DET-COMPASS dataset builds upon COMPASS-XP by extending the annotations with manually labeled bounding boxes (\cref{fig:compass_qualitative}). The annotation process was conducted by hiring three experts, each responsible for labeling 50\% of the RGB-X-ray pairs. To ensure accurate alignment between the RGB and X-ray images, each expert annotated both modalities simultaneously. After completing their respective sets, all three experts reviewed the annotations collectively. One of them acted as a middle ground, overseeing the review process and resolving any remaining discrepancies to ensure annotation consistency.

In total, DET-COMPASS comprises 3,856 annotated images, including 1,928 X-ray and 1,928 RGB images. The average annotation time per image, regardless of modality, was 20 seconds. Given that each expert annotated half of the dataset, the total annotation time amounted to 32.13 hours. The review process required an additional 3 seconds per image, and since all experts participated in reviewing the entire dataset, the total review time was 9.64 hours.

The total number of annotated objects (bounding boxes) in the X-ray images is 1,907, while in the RGB images, it is 1,870. This discrepancy arises because some objects are occluded in the RGB modality, making their localization impossible (\cref{fig:compass_qualitative_p1}). Each annotated object in the X-ray modality includes a \textit{visibility} attribute, indicating whether it produces a discernible signature in the X-ray spectrum. An example of an object marked as visible is shown in \cref{fig:compass_qualitative_p2}\hyperref[fig:compass_qualitative_p2]{(2)}, while an example of an object marked as non-visible is presented in \cref{fig:compass_qualitative_p2}\hyperref[fig:compass_qualitative_p2]{(1)}. DET-COMPASS comprises a total of 370 object classes (detailed in \cref{tab:categories_compass}), of which 307 contain at least one annotated visible object.

Finally, DET-COMPASS avoids long-tail distribution issues thanks to its uniformly distributed categories, with a low Gini coefficient of $G = 0.26$ (\eg, MS-COCO has $G = 0.57$, where higher $G$ indicates bigger long-tail bias).

\section{Further Implementation Details of \method}
\label{sec:details2}

\subsection{Pseudo-code of \method}

In \cref{algorithim:extended}, we present the pseudocode for the core implementation of \method, detailing both the construction of visual descriptors and their use to classify detector proposals. 

\subsection{Material-Transfer Mechanism}

To construct the material database $\mathcal{M}$, we cluster $\mathcal{C^{\text{in-house}}}$ into groups of materials identified by a large language model (LLM). The average appearance of objects within each group is used as an estimator of the corresponding material. To perform this clustering, we utilize GPT-4 with the prompt specified in \cref{tab:promts}\hyperref[tab:promts]{(1)}.

Once the material database is computed, it can be used to adapt RGB objects to the X-ray modality by inpainting them with their expected material. These expected materials are retrieved from $\mathcal{M}$ using an LLM with the prompt provided in \cref{tab:promts}\hyperref[tab:promts]{(2)}.

\paragraph{Material database construction when $\mathcal{D}^{\text{in-house}}_{\text{XRAY}} = \emptyset$.} 

When no samples are available from $\mathcal{D}^{\text{in-house}}_{\text{XRAY}}$, we construct our material database using the standardized color scheme of security X-ray scans. These scans operate by irradiating objects with X-rays and rendering them in pseudo-colors based on their spectral absorption rates. Typically, three primary pseudo-colors are used~\cite{akcay2022towards,velayudhan2022recent}: \textbf{orange} for organic substances (\eg, food, explosives), \textbf{green} for inorganic materials (\eg, laptops, smartphones), and \textbf{blue} for metals (\eg, knives, guns). We leverage this modality knowledge to build our material database around these three broad materials.

\subsection{Web-retrieval Details}
To retrieve images from the web, we utilize the Google Custom Search API~\cite{googleapi}, configuring specific query parameters to refine the results. We set the search type to images (\texttt{searchType:} \texttt{image}) and restrict the results to photos (\texttt{imgType:} \texttt{photo}) in common JPEG and PNG formats (\texttt{fileType:} \texttt{jpg|png}). To ensure relevance, we limit searches to English-language sources (\texttt{lr:} \texttt{lang\_en}) and prioritize images from the past seven years (\texttt{dateRestrict:} \texttt{y7}). 

\subsection{In-domain Descriptor Details}

In-domain descriptors from $\mathcal{D}^{\text{in-house}}_{\text{XRAY}}$ are built offline by combining the training sets from the six evaluation datasets (PIXray~\cite{pixray}, PIDray~\cite{wang2021towards}, CLCXray~\cite{clcxray}, DvXray~\cite{dvxray}, HiXray~\cite{HiXray}, and DET-COMPASS) and removing overlapping categories. Combining the datasets ensures a fair evaluation through dataset-agnostic prototypes that capture generic concepts, rather than dataset-specific representations.

\begin{algorithm}[!t]
\SetAlgoLined
\DontPrintSemicolon
\SetNoFillComment
\footnotesize

\KwIn{vocabulary $\classTest$; OvOD detector $\mathcal{F}$; test image $\mathbf{I}$; in-house database $\mathcal{D}^{\text{in-house}}_{\text{XRAY}}$; web-database $\mathcal{D}^{\text{web}}_{\text{RGB}}$ }

\KwOut{Detections $\mathcal{T}$ of image  $\mathbf{I}$}
\BlankLine

Initialization: $\mathcal{T} \leftarrow \emptyset$\; 
Initialization: $\mathcal{X} \leftarrow \emptyset$\; 
Initialization: $\mathcal{X}_{bg}\leftarrow \emptyset$\;
\BlankLine

$\mathcal{M} = CreateMaterialDatabase(\mathcal{D}^{\text{in-house}}_{\text{XRAY}})$ \;
\BlankLine

\tcc{Visual class descriptors construction}
\BlankLine

\For{\text{class} $c \in \classTest$}{
    
    \BlankLine
    \tcc{VSA refers to the Visual samples acquisition pipeline}
    \BlankLine

    $\mathcal{G}_{c}^{\text{XRAY}} \leftarrow VSA(c, \mathcal{D}^{\text{in-house}}_{\text{XRAY}})$ \;
    \If{$\mathcal{G}_{c}^{\text{XRAY}}$ is $\emptyset$}{
        $\widetilde{\mathcal{G}}_{c}^{\text{web}} \leftarrow VSA(c, \mathcal{D}^{\text{web}}_{\text{RGB}})$ \;
        $\mathcal{G}_{c}^{\text{web}} = \operatorname{Filter}(\widetilde{\mathcal{G}}_{c}^{\text{web}}, \mathcal{F}, c, \tau)$\;
        $\mathcal{A}_m^c = GetMaterialAppareance(\mathcal{M}, c)$ \;
        \For{sample $\mathbf{u} \in \mathcal{G}_{c}^{\text{web}}$}{
            \tcc{$\Omega$ denotes segmentation}
            $\tilde{\mathbf{u}} = \Omega(\mathbf{u}) \odot (\mathcal{A}_m^c \cdot \mathbf{1})$\;
            $\mathcal{G}_c^{\text{XRAY}} \leftarrow \mathcal{G}_c^{\text{XRAY}} \cup \{\tilde{\mathbf{u}}\} $ \;
        }
    }

    \BlankLine
    \tcc{Visual class modeling}
    \BlankLine

    $\mathcal{X}_c \leftarrow \emptyset$ \;
    \For{sample $\mathbf{I} \in \mathcal{G}_c^{\text{XRAY}}$}{

        $\mathbf{x}^{\text{pos}}_\mathbf{I} = \cref{eq:pos_prot}$ \;
        
        $\mathbf{x}^{\text{neg}}_\mathbf{I} = \cref{eq:neg_prot}$ \;
        
        $\mathcal{X}_c \leftarrow \mathcal{X}_c \cup \{\mathbf{x}^{\text{pos}}_I$\} \;
        $\mathcal{X}_{bg} \leftarrow \mathcal{X}_{bg} \cup \{\mathbf{x}^{\text{neg}}_I$\} \;
    }
     $\mathcal{X}_c \leftarrow \mathcal{X}_c \cup \{Avg(\mathcal{X}_c)\}$ \;
     $\mathcal{X} \leftarrow \mathcal{X} \cup \mathcal{X}_c$

}

\BlankLine
\tcc{Detection on image I}
\BlankLine

$z = \mathcal{F} \mid \Phi_{RPN}(\mathbf{I})$\;
$\mathcal{C}^{\text{test}^\prime} \leftarrow \classTest \cup \{\texttt{background}\}$ \;

\BlankLine
\For{proposal $\mathbf{z_m} \in z$}{
    $\hat{c_m} \leftarrow \argmax_{c \in \mathcal{C}^{\text{test}^\prime}} \max_{\mathcal{X}_c^i \in \mathcal{X}_c} \langle \mathbf{z}_m, \mathcal{X}_c^i \rangle$ \;
    
    $\hat{\mathbf{b_m}} \leftarrow \mathcal{F} \mid \Phi_{REG} (\mathbf{z_m})$ \;
    \BlankLine

    \tcc{DCC refers to the Descriptor Consistency Criterion}
    \BlankLine

    \If{$\hat{c_m}$ is not $\texttt{background}$ and $DCC(\mathbf{z_m},\mathcal{X})$}{
        $\mathcal{T} \leftarrow \{\hat{c_m} \cup \hat{\mathbf{b_m}}\}$
    }
}

Return: $\mathcal{T}$
\caption{Pseudo-code of \method.}
\label{algorithim:extended}
\end{algorithm}

\subsection{Dataset Colorization}
We do not perform color adjustments across datasets, as most do not provide raw density values.
However, this does not adversely affect \method, since the colorization strategies follow manufacturer-specific yet \emph{consistent palettes} that use similar colors to represent the same materials.
These mappings, while differing slightly in hue or intensity, consistently represent the material-specific density and spatial structure necessary for robust detection.
Notably, our DET-COMPASS also includes raw density values, enabling more flexible experimentation in future work.

\subsection{Complexity Analysis}
\method is designed to adapt \emph{off-the-shelf} RGB OvOD methods to X-ray without training, making it inherently \emph{modular}.
Importantly, most of its components run \emph{offline} \emph{only once} to build the visual descriptors, requiring roughly 0.7s per class on an NVIDIA A100 GPU.
At inference, \method simply replaces the text-based classifier of the base OvOD detector with its visual-based classifier, introducing negligible overhead (\eg, 3ms/sample on G-DINO) with complexity $\bigO{n}$ \wrt the number of categories.

\section{Extended Experimental Results}
\label{sec:results2}

Maintaining the same experimental setup as in \cref{sec:ovod_results}, we extend our main results to report AP, AP50, and AP75. Additionally, since the experiments are repeated three times with different random distributions of in-domain and web categories for the intermediate gallery settings, we also report the standard deviation. \cref{tab:main_exps_extended} show the results. The low standard deviations, combined with RAXO’s consistent improvement over all baselines, further validate the effectiveness of \method in adapting \textit{off-the-shelf} open-vocabulary detectors to the X-ray modality.

To validate \method with an LLM-guided DETR, we also integrated it into LaMI-DETR~\cite{du2024lami}, yielding consistent improvements across all settings (\cref{tab:main_exps_lami}). Finally, to show that the large models in \method can be removed or replaced to achieve a desired balance between efficiency and precision, we present an additional ablation in \cref{tab:comp_ablation}.

\begin{table}[!t]
\setlength\extrarowheight{-2pt}
\centering
\resizebox{0.80\columnwidth}{!}{%
\begin{tabular}{l|c@{\hskip 8pt}c}
\toprule
\textbf{Category} & \textbf{G-DINO} & \textbf{G-DINO+RAXO} \\
\midrule
Pressure Vessel   & 0.5  & \textbf{52.8} \gooddelta{52.3} \\
Bat               & 70.7 & 69.7 \baddelta{-1.0} \\
Gun               & 31.3 & \textbf{53.6} \gooddelta{22.3} \\
Scissors          & 29.6 & \textbf{44.3} \gooddelta{14.7} \\
Razor Blade       & 0.9  & \textbf{18.1} \gooddelta{17.2} \\
Pliers            & 12.4 & \textbf{43.5} \gooddelta{31.1} \\
Dart              & 0.4  & \textbf{32.0} \gooddelta{31.6} \\
Knife             & 6.2  & \textbf{10.3} \gooddelta{4.1} \\
Fireworks         & 0.0  & \textbf{2.1}  \gooddelta{2.1} \\
Battery           & 5.9  & \textbf{47.7} \gooddelta{41.8} \\
Saw Blade         & 3.2  & \textbf{23.9} \gooddelta{20.7} \\
Hammer            & 1.3  & \textbf{56.1} \gooddelta{54.8} \\
Screwdriver       & 1.0  & \textbf{19.9} \gooddelta{18.9} \\
Wrench            & 28.2 & \textbf{52.3} \gooddelta{24.1} \\
Lighter           & 2.0  & \textbf{26.8} \gooddelta{24.8} \\
\midrule
\textbf{Average}  & 12.9 & \textbf{36.9} \gooddelta{+24.0} \\
\bottomrule
\end{tabular}
}
\caption{\textbf{Per-category AP comparison} on the PIXray~\cite{pixray} dataset for G-DINO~\cite{liu2025grounding}. \method significantly boosts performance across nearly all categories, particularly those with low G-DINO baseline scores.}
\label{tab:category_ap}
\end{table}

\begin{table}[!t]
\setlength\extrarowheight{-2pt}
\centering
\resizebox{1.0\columnwidth}{!}{%
\begin{tabular}{@{}ll@{\hskip 0pt}|c@{\hskip 2pt}c@{\hskip 2pt}c@{\hskip 2pt}c@{\hskip 2pt}c@{\hskip 4pt}|c@{\hskip 2pt}c@{\hskip 2pt}c@{\hskip 2pt}c@{\hskip 2pt}c@{\hskip 2pt}@{\hskip -2pt}}
\toprule

& & %
\multirow{2}{*}{
    \makecell{
    \rotatebox[origin=c]{30}{\hspace{-20pt}{\smaller[0.0] \textbf{Binder\hspace{-5pt}}}}
    }%
  }%
& %
\multirow{2}{*}{
    \makecell{
    \rotatebox[origin=c]{30}{\hspace{-15pt}{\smaller[0.0] \textbf{Milk\hspace{-0pt}}}} \\[-2pt]
    \rotatebox[origin=c]{30}{\hspace{-15pt}{\smaller[0.0] \textbf{carton\hspace{-5pt}}}}
    }%
  }%
& %
\multirow{2}{*}{
    \makecell{
    \rotatebox[origin=c]{30}{\hspace{-15pt}{\smaller[0.0] \textbf{Crayon\hspace{-5pt}}}}
    }%
  }%
& %
\multirow{2}{*}{
    \makecell{
    \rotatebox[origin=c]{30}{\hspace{-15pt}{\smaller[0.0] \textbf{Hair\hspace{-0pt}}}} \\[-2pt]
    \rotatebox[origin=c]{30}{\hspace{-15pt}{\smaller[0.0] \textbf{gel\hspace{-5pt}}}}
    }%
  }%
& %
\multirow{2}{*}{
    \makecell{
    \rotatebox[origin=c]{30}{\hspace{-15pt}{\smaller[0.0] \textbf{Crowbar\hspace{-5pt}}}}
    }%
  }%
& %
\multirow{2}{*}{
    \makecell{
    \rotatebox[origin=c]{30}{\hspace{-15pt}{\smaller[0.0] \textbf{Can\hspace{-0pt}}}} \\[-4pt]
    \rotatebox[origin=c]{30}{\hspace{-15pt}{\smaller[0.0] \textbf{opener\hspace{-5pt}}}}
    }%
  }%
& %
\multirow{2}{*}{
    \makecell{
    \rotatebox[origin=c]{30}{\hspace{-20pt}{\smaller[0.0] \textbf{Corkscrew\hspace{-5pt}}}}
    }%
  }%
& %
\multirow{2}{*}{
    \makecell{
    \rotatebox[origin=c]{30}{\hspace{-15pt}{\smaller[0.0] \textbf{Strainer\hspace{-5pt}}}}
    }%
  }%
& %
\multirow{2}{*}{
    \makecell{
    \rotatebox[origin=c]{30}{\hspace{-15pt}{\smaller[0.0] \textbf{High\hspace{-0pt}}}} \\[-1pt]
    \rotatebox[origin=c]{30}{\hspace{-15pt}{\smaller[0.0] \textbf{heel\hspace{-5pt}}}}
    }%
  }%
& %
\multirow{2}{*}{
    \makecell{
    \rotatebox[origin=c]{30}{\hspace{-20pt}{\smaller[1.0] \textbf{Compact\hspace{-0pt}}}} \\[-2pt]
    \rotatebox[origin=c]{30}{\hspace{-20pt}{\smaller[0.0] \textbf{disc\hspace{-5pt}}}}
    }%
  }%
\\[16pt]

\midrule
\midrule

\multicolumn{2}{@{}l@{\hskip 2pt}|}{G-DINO} & 0.0 & 0.1 & 0.1 & 0.2 & 3.0 & 7.5 & 1.0 & 16.7 & 14.3 & 1.3 \\[-3pt]

\midrule

\plusours\hspace{-10pt} & & \textbf{0.7}\gooddelta{0.7} & \textbf{1.3}\gooddelta{1.2} & \textbf{1.3}\gooddelta{1.2} & \textbf{2.6}\gooddelta{2.4} & \textbf{4.2}\gooddelta{1.1} & \textbf{88.9}\gooddelta{81.4} & \textbf{90.1}\gooddelta{89.1} & \textbf{98.2}\gooddelta{81.5} & \textbf{98.9}\gooddelta{84.6} & \textbf{99.1}\gooddelta{97.8}\\ [-1pt]

\bottomrule
\end{tabular}
}

\caption{\textbf{Per-category AP on DET-COMPASS} for the top-5 classes with the highest and lowest performance gains. We report AP for G-DINO~\cite{liu2025grounding} and G-DINO+\method across categories.
}

\label{tab:per_cat_ap_compass}

\end{table}

\begin{table*}[!t]
\centering
\begin{adjustbox}{width=\textwidth}
\begin{tabular}{@{}c@{\hskip 4pt}c@{\hskip -2pt}l@{\hskip 2pt}|c@{\hskip 4pt}c@{\hskip 4pt}c@{\hskip 2pt}|c@{\hskip 4pt}c@{\hskip 4pt}c@{\hskip 2pt}|c@{\hskip 4pt}c@{\hskip 4pt}c@{\hskip 2pt}|c@{\hskip 4pt}c@{\hskip 4pt}c@{\hskip 2pt}|c@{\hskip 4pt}c@{\hskip 4pt}c@{\hskip 2pt}|c@{\hskip 4pt}c@{\hskip 4pt}c@{}}
\toprule
\multicolumn{2}{c}{$\bm{\mathcal{G}}$} & Method & \multicolumn{3}{c}{\textbf{PIXRAY}} & \multicolumn{3}{c}{\textbf{PIDRAY}} & \multicolumn{3}{c}{\textbf{CLCXray}} & \multicolumn{3}{c}{\textbf{COMPASS-XP}} &  \multicolumn{3}{c}{\textbf{HiXray}} & \multicolumn{3}{c}{\textbf{DVXray}} \\
& & & AP & AP50 & AP75 & AP & AP50 & AP75 & AP & AP50 & AP75 & AP & AP50 & AP75 & AP & AP50 & AP75 & AP & AP50 & AP75 \\
\midrule
\midrule
&&G-DINO~\cite{liu2025grounding}& 12.9 & 14.9 & 13.4 & 10.9 & 13.6 & 11.7 & 6.7 & 8.4 & 7.1 & 13.4 & 14.2 & 13.9   & 7.0 & 10.8 & 8.2 & 10.0 & 11.2 & 10.4 \\
\midrule
\multirow{5}{*}{
    \makecell{
    \rotatebox[origin=c]{45}{$\mathcal{D}^{\text{in-h}}_{\text{{\smaller{XRAY\hspace{-50pt}}}}}$}\\[-5pt]
    \tikz[baseline=(current bounding box.center)]{
      \draw[->, thick] (0, 0.6) -- (0, 0.1); 
    }\\[-2pt]
    \rotatebox[origin=c]{45}{$\mathcal{D}^{\text{web}}_{\text{{\smaller{RGB}}}}$}
    }%
  }%
& 100/0& & 36.9 & 45.0 & 39.0 & 16.5 & 21.4 & 17.9 & 22.2 & 29.6 & 24.4 & 47.9 & 54.2 & 48.8   & 17.1 & 27.2 & 19.4 & 22.6 & 26.6 & 24.1 \\
& 80/20& & 33.8$\pm${\smaller{0.6}} & 40.9$\pm${\smaller{0.9}} & 35.5$\pm${\smaller{0.6}} & 15.4$\pm${\smaller{0.4}} & 19.8$\pm${\smaller{0.6}} & 16.6$\pm${\smaller{0.4}} & 18.0$\pm${\smaller{2.1}} & 23.7$\pm${\smaller{2.3}} & 19.5$\pm${\smaller{2.2}} & 41.0$\pm${\smaller{2.2}} & 46.2$\pm${\smaller{2.4}} & 41.7$\pm${\smaller{2.2}} & 14.5$\pm${\smaller{0.6}} & 23.5$\pm${\smaller{1.0}} & 16.3$\pm${\smaller{0.6}} & 21.0$\pm${\smaller{0.6}} & 24.8$\pm${\smaller{0.9}} & 22.3$\pm${\smaller{0.6}} \\
& 50/50& \plusours  & 25.4$\pm${\smaller{2.0}} & 31.2$\pm${\smaller{1.9}} & 26.7$\pm${\smaller{2.0}} & 15.5$\pm${\smaller{0.9}} & 19.8$\pm${\smaller{1.0}} & 16.8$\pm${\smaller{1.0}} & 17.0$\pm${\smaller{1.8}} & 22.9$\pm${\smaller{3.2}} & 18.7$\pm${\smaller{2.3}} & 31.4$\pm${\smaller{0.7}} & 35.3$\pm${\smaller{0.9}} & 32.1$\pm${\smaller{0.6}}   & 13.4$\pm${\smaller{0.1}} & 21.3$\pm${\smaller{0.1}} & 15.3$\pm${\smaller{0.2}} & 16.1$\pm${\smaller{1.8}} & 18.8$\pm${\smaller{2.3}} & 17.0$\pm${\smaller{2.0}} \\
& 20/80& & 21.6$\pm${\smaller{0.6}} & 26.1$\pm${\smaller{1.1}} & 22.6$\pm${\smaller{0.6}} & 13.9$\pm${\smaller{0.5}} & 17.9$\pm${\smaller{0.7}} & 14.9$\pm${\smaller{0.6}} & 10.0$\pm${\smaller{0.4}} & 13.1$\pm${\smaller{1.5}} & 10.7$\pm${\smaller{0.7}} & 20.5$\pm${\smaller{0.6}} & 22.9$\pm${\smaller{0.7}} & 21.1$\pm${\smaller{0.7}}   & 9.8$\pm${\smaller{1.0}} & 15.8$\pm${\smaller{1.4}} & 11.1$\pm${\smaller{1.2}} & 15.0$\pm${\smaller{1.0}} & 17.2$\pm${\smaller{1.1}} & 15.8$\pm${\smaller{1.2}} \\
& 0/100& & 16.1 & 19.8 & 16.8 & 13.4 & 17.1 & 14.3 & 7.1 & 9.7 & 7.5 & 14.0 & 15.4 & 14.5   & 7.9 & 13.0 & 8.7 & 12.4 & 14.1 & 12.9 \\
\midrule
\midrule
& & Detic~\cite{detic} & 9.3 & 11.6 & 9.5 & 7.1 & 9.7 & 7.6 & 4.7 & 7.3 & 4.6 & 11.5 & 13.4 & 13.3   & 4.8 & 8.6 & 5.2 & 7.0 & 8.5 & 7.5 \\
\midrule
\multirow{5}{*}{
    \makecell{
    \rotatebox[origin=c]{45}{$\mathcal{D}^{\text{in-h}}_{\text{{\smaller{XRAY\hspace{-50pt}}}}}$}\\[-5pt]
    \tikz[baseline=(current bounding box.center)]{
      \draw[->, thick] (0, 0.6) -- (0, 0.1); 
    }\\[-2pt]
    \rotatebox[origin=c]{45}{$\mathcal{D}^{\text{web}}_{\text{{\smaller{RGB}}}}$}
    }%
  }%
& 100/0& & 27.3 & 34.5 & 28.2 & 11.3 & 15.8 & 12.2 & 14.0 & 20.6 & 14.7 & 35.3 & 39.9 & 35.4   & 14.2 & 23.9 & 15.5 & 19.4 & 23.9 & 21.2 \\
& 80/20& & 23.9$\pm${\smaller{1.3}} & 30.2$\pm${\smaller{1.5}} & 24.6$\pm${\smaller{1.3}} & 10.8$\pm${\smaller{0.1}} & 15.0$\pm${\smaller{0.2}} & 11.7$\pm${\smaller{0.1}} & 12.3$\pm${\smaller{1.6}} & 18.1$\pm${\smaller{1.8}} & 12.8$\pm${\smaller{1.9}} & 30.7$\pm${\smaller{1.4}} & 34.4$\pm${\smaller{1.3}} & 30.8$\pm${\smaller{1.5}} & 12.1$\pm${\smaller{1.1}} & 20.8$\pm${\smaller{1.8}} & 13.1$\pm${\smaller{1.2}} & 18.0$\pm${\smaller{2.2}} & 22.1$\pm${\smaller{2.6}} & 19.7$\pm${\smaller{2.4}} \\
& 50/50&\plusours  & 19.5$\pm${\smaller{1.6}} & 24.8$\pm${\smaller{1.9}} & 20.1$\pm${\smaller{1.7}} & 10.3$\pm${\smaller{0.3}} & 14.3$\pm${\smaller{0.3}} & 11.0$\pm${\smaller{0.3}} & 9.2$\pm${\smaller{1.2}} & 13.5$\pm${\smaller{2.3}} & 9.5$\pm${\smaller{1.2}} & 24.4$\pm${\smaller{2.7}} & 27.1$\pm${\smaller{2.7}} & 24.8$\pm${\smaller{2.6}}   & 11.0$\pm${\smaller{0.9}} & 18.9$\pm${\smaller{1.3}} & 11.9$\pm${\smaller{1.2}} & 14.6$\pm${\smaller{1.1}} & 17.9$\pm${\smaller{1.2}} & 15.9$\pm${\smaller{1.2}} \\
& 20/80& & 15.2$\pm${\smaller{0.9}} & 19.4$\pm${\smaller{0.9}} & 15.5$\pm${\smaller{1.0}} & 9.6$\pm${\smaller{0.1}} & 13.3$\pm${\smaller{0.2}} & 10.3$\pm${\smaller{0.2}} & 8.0$\pm${\smaller{0.1}} & 12.5$\pm${\smaller{0.1}} & 8.0 & 16.4$\pm${\smaller{1.0}} & 18.3$\pm${\smaller{1.0}} & 16.4$\pm${\smaller{1.0}}   & 9.9$\pm${\smaller{0.8}} & 16.8$\pm${\smaller{1.4}} & 10.7$\pm${\smaller{0.9}} & 12.7$\pm${\smaller{0.6}} & 15.5$\pm${\smaller{0.8}} & 13.9$\pm${\smaller{0.7}} \\
& 0/100& & 13.4 & 16.8 & 13.6 & 9.1 & 12.6 & 9.8 & 5.2 & 8.1 & 5.1 & 11.9 & 13.1 & 12.1 & 7.9 & 13.8 & 8.4 & 9.4 & 11.4 & 10.1 \\
\midrule
\midrule
& & CoDet~\cite{codet} & 7.3 & 8.7 & 7.6 & 5.7 & 7.6 & 6.2 & 3.1 & 5.7 & 2.7 & 8.4 & 8.9 & 8.7   & 3.4 & 5.9 & 3.7 & 5.6 & 6.8 & 6.0 \\
\midrule
\multirow{5}{*}{
    \makecell{
    \rotatebox[origin=c]{45}{$\mathcal{D}^{\text{in-h}}_{\text{{\smaller{XRAY\hspace{-50pt}}}}}$}\\[-5pt]
    \tikz[baseline=(current bounding box.center)]{
      \draw[->, thick] (0, 0.6) -- (0, 0.1); 
    }\\[-2pt]
    \rotatebox[origin=c]{45}{$\mathcal{D}^{\text{web}}_{\text{{\smaller{RGB}}}}$}
    }%
  }%
& 100/0 & & 27.9 & 33.6 & 29.2 & 10.3 & 14.6 & 10.9 & 14.8 & 22.4 & 15.9 & 35.8 & 41.0 & 36.7   & 13.2 & 22.0 & 14.8 & 17.6 & 21.7 & 19.0 \\
& 80/20 & & 25.1$\pm${\smaller{1.5}} & 30.2$\pm${\smaller{1.7}} & 26.2$\pm${\smaller{1.7}} & 9.5$\pm${\smaller{0.3}} & 13.4$\pm${\smaller{0.5}} & 10.1$\pm${\smaller{0.3}} & 12.0$\pm${\smaller{1.9}} & 18.3$\pm${\smaller{2.8}} & 12.7$\pm${\smaller{2.1}} & 32.2$\pm${\smaller{0.9}} & 36.5$\pm${\smaller{1.5}} & 33.1$\pm${\smaller{0.6}}   & 11.7$\pm${\smaller{1.3}} & 19.4$\pm${\smaller{2.2}} & 13.2$\pm${\smaller{1.5}} & 15.4$\pm${\smaller{1.4}} & 18.8$\pm${\smaller{1.7}} & 16.7$\pm${\smaller{1.6}} \\
& 50/50 &\plusours  & 20.0$\pm${\smaller{0.7}} & 24.1$\pm${\smaller{0.9}} & 20.8$\pm${\smaller{0.7}} & 9.5$\pm${\smaller{0.5}} & 13.4$\pm${\smaller{0.7}} & 10.1$\pm${\smaller{0.5}} & 9.2$\pm${\smaller{1.4}} & 14.2$\pm${\smaller{2.1}} & 9.6$\pm${\smaller{1.7}} & 24.0$\pm${\smaller{0.2}} & 26.7$\pm${\smaller{0.3}} & 24.7$\pm${\smaller{0.2}}   & 9.9$\pm${\smaller{0.4}} & 16.7$\pm${\smaller{0.8}} & 11.1$\pm${\smaller{0.4}} & 11.5$\pm${\smaller{0.8}} & 14.2$\pm${\smaller{1.1}} & 12.4$\pm${\smaller{0.8}} \\
& 20/80 & & 14.8$\pm${\smaller{2.4}} & 17.8$\pm${\smaller{2.8}} & 15.3$\pm${\smaller{2.5}} & 8.5$\pm${\smaller{0.3}} & 11.9$\pm${\smaller{0.4}} & 9.0$\pm${\smaller{0.4}} & 5.1$\pm${\smaller{1.4}} & 9.0$\pm${\smaller{2.5}} & 5.0$\pm${\smaller{1.6}} & 17.8$\pm${\smaller{0.7}} & 19.4$\pm${\smaller{0.9}} & 18.2$\pm${\smaller{0.6}}   & 8.1$\pm${\smaller{0.6}} & 13.8$\pm${\smaller{1.0}} & 8.8$\pm${\smaller{0.6}} & 9.4$\pm${\smaller{1.5}} & 11.3$\pm${\smaller{1.8}} & 10.1$\pm${\smaller{1.6}} \\
& 0/100 & & 11.5 & 14.0 & 11.9 & 8.1 & 11.3 & 8.7 & 4.0 & 7.1 & 3.8 & 12.2 & 13.0 & 12.6  & 6.5 & 11.2 & 7.1 & 6.9 & 8.3 & 7.5 \\
\midrule
\midrule
& & VLDet~\cite{vldet} & 9.8 & 12.1 & 10.3 & 6.9 & 9.4 & 7.4 & 4.4 & 7.8 & 4.0 & 10.6 & 11.4 & 10.8   & 5.1 & 9.0 & 5.5 & 7.4 & 9.2 & 8.1 \\
\midrule
\multirow{5}{*}{
    \makecell{
    \rotatebox[origin=c]{45}{$\mathcal{D}^{\text{in-h}}_{\text{{\smaller{XRAY\hspace{-50pt}}}}}$}\\[-5pt]
    \tikz[baseline=(current bounding box.center)]{
      \draw[->, thick] (0, 0.6) -- (0, 0.1); 
    }\\[-2pt]
    \rotatebox[origin=c]{45}{$\mathcal{D}^{\text{web}}_{\text{{\smaller{RGB}}}}$}
    }%
  }%
& 100/0 & & 32.3 & 40.1 & 34.0 & 11.7 & 16.8 & 12.6 & 15.4 & 23.3 & 15.9 & 36.4 & 41.4 & 37.2   & 14.8 & 24.5 & 16.3 & 20.1 & 25.1 & 22.0 \\
& 80/20 & & 29.2$\pm${\smaller{1.2}} & 36.3$\pm${\smaller{1.2}} & 30.7$\pm${\smaller{1.3}} & 11.0$\pm${\smaller{0.3}} & 15.7$\pm${\smaller{0.3}} & 11.7$\pm${\smaller{0.3}} & 12.7$\pm${\smaller{0.5}} & 19.6$\pm${\smaller{1.2}} & 13.0$\pm${\smaller{0.5}} & 31.8$\pm${\smaller{0.8}} & 36.0$\pm${\smaller{1.0}} & 32.5$\pm${\smaller{0.9}}   & 13.1$\pm${\smaller{1.2}} & 21.8$\pm${\smaller{1.9}} & 14.3$\pm${\smaller{1.3}} & 16.8$\pm${\smaller{0.2}} & 21.0$\pm${\smaller{0.1}} & 18.4$\pm${\smaller{0.1}} \\
& 50/50 &\plusours  & 24.0$\pm${\smaller{1.5}} & 29.9$\pm${\smaller{1.7}} & 25.2$\pm${\smaller{1.5}} & 10.4$\pm${\smaller{0.7}} & 14.6$\pm${\smaller{1.0}} & 11.1$\pm${\smaller{0.8}} & 11.1$\pm${\smaller{1.1}} & 16.9$\pm${\smaller{0.4}} & 11.5$\pm${\smaller{1.7}} & 23.7$\pm${\smaller{0.9}} & 26.5$\pm${\smaller{0.8}} & 24.3$\pm${\smaller{1.1}}   & 11.2$\pm${\smaller{1.5}} & 19.0$\pm${\smaller{2.1}} & 12.1$\pm${\smaller{1.9}} & 12.1$\pm${\smaller{0.5}} & 15.0$\pm${\smaller{0.4}} & 13.2$\pm${\smaller{0.4}} \\
& 20/80 & & 21.6$\pm${\smaller{1.0}} & 26.8$\pm${\smaller{0.9}} & 22.6$\pm${\smaller{1.0}} & 9.4$\pm${\smaller{0.3}} & 13.3$\pm${\smaller{0.4}} & 10.1$\pm${\smaller{0.3}} & 5.2$\pm${\smaller{0.1}} & 9.1$\pm${\smaller{0.2}} & 4.8$\pm${\smaller{0.0}} & 16.2$\pm${\smaller{0.9}} & 18.2$\pm${\smaller{1.2}} & 16.6$\pm${\smaller{1.0}}   & 9.3$\pm${\smaller{0.2}} & 15.9$\pm${\smaller{0.2}} & 9.9$\pm${\smaller{0.3}} & 10.6$\pm${\smaller{0.5}} & 13.1$\pm${\smaller{0.6}} & 11.5$\pm${\smaller{0.5}} \\
& 0/100 & & 14.1 & 17.8 & 14.5 & 8.9 & 12.5 & 9.5 & 4.4 & 8.1 & 3.9 & 11.1 & 12.2 & 11.4   & 8.3 & 14.5 & 8.7 & 9.0 & 11.0 & 9.8 \\
\bottomrule
\end{tabular}
\end{adjustbox}
\caption{
\textbf{X-ray OvOD performance under the Cross-Modality Transfer Evaluation (CMTE) setting} on DET-COMPASS (ours), PIXray~\cite{pixray}, PIDray~\cite{wang2021towards}, CLCXray~\cite{clcxray}, DvXray~\cite{dvxray}, and HiXray~\cite{HiXray} datasets.
We integrate \method into different baselines using different gallery $\mathcal{G}$ compositions, from using only $\mathcal{D}^{\text{in-house}}_{\text{XRAY}}$ data (100/0) to exclusively $\mathcal{D}^{web}_{\text{RGB}}$ samples (0/100).
\method consistently improves the performance of all baseline OvOD detectors across every dataset. We report the AP, AP50 and AP75. We also include the deviations because each experiment is repeated three times with different random distributions of in-domain and web categories for the intermediate gallery settings.
}
\label{tab:main_exps_extended}
\end{table*}

\begin{table}[t]
\setlength\extrarowheight{-3pt}
\centering
\resizebox{1.0\linewidth}{!}{%
\begin{tabular}{@{}c@{\hskip 6pt}l|U@{\hskip 0.5pt}L@{\hskip 0pt}R@{\hskip 0.5pt}L@{\hskip 0pt}R@{\hskip 0.5pt}L@{\hskip 0pt}R@{\hskip 0.5pt}L@{\hskip 0pt}R@{\hskip 0.5pt}L@{\hskip 0pt}R@{\hskip 0.5pt}Ll@{\hskip -15pt}}
\toprule

$\bm{\mathcal{G}}$ & \multicolumn{1}{c|}{\textbf{Method}} & \multicolumn{2}{c}{\textbf{D-COMP.}} & \multicolumn{2}{c}{\textbf{PIXray}} & \multicolumn{2}{c}{\textbf{PIDray}} & \multicolumn{2}{c}{\textbf{CLCXray}} & \multicolumn{2}{c}{\textbf{DvXray}} & \multicolumn{2}{c}{\textbf{HiXray}} \\

\midrule
\midrule

\multicolumn{2}{r|}{LaMI-DETR} & \multicolumn{2}{c}{11.3} & \multicolumn{2}{c}{13.6} & \multicolumn{2}{c}{8.0} & \multicolumn{2}{c}{4.0} & \multicolumn{2}{c}{9.7} & \multicolumn{2}{c}{6.3} \\

\midrule

\textit{100/0} & & \textbf{31.9} & \gooddelta{20.6} & \textbf{25.7} & \gooddelta{12.1} & \textbf{13.1} & \gooddelta{5.1} & \textbf{18.7} & \gooddelta{14.7} & \textbf{18.1} & \gooddelta{8.4} & \textbf{9.7} & \gooddelta{3.4} &\\

\textit{80/20} & & \textbf{27.2} & \gooddelta{15.9} & \textbf{23.2} & \gooddelta{9.6} & \textbf{12.0} & \gooddelta{4.0} & \textbf{16.2} & \gooddelta{12.2} & \textbf{16.0} & \gooddelta{6.3} & \textbf{8.3} & \gooddelta{2.0} &\\

\textit{50/50} & \plusours & \textbf{22.0} & \gooddelta{10.7} & \textbf{15.9} & \gooddelta{2.3} & \textbf{12.1} & \gooddelta{4.1} & \textbf{15.1} & \gooddelta{11.1} & \textbf{12.8} & \gooddelta{3.1} & \textbf{7.6} & \gooddelta{1.3} &\\

\textit{20/80} & & \textbf{15.7} & \gooddelta{4.4} & \textbf{15.2} & \gooddelta{1.6} & \textbf{10.8} & \gooddelta{2.8} & \textbf{6.8} & \gooddelta{2.8} & \textbf{11.9} & \gooddelta{2.2} & \textbf{7.0} & \gooddelta{0.7} &\\

\textit{0/100} & & \textbf{11.5} & \gooddelta{0.2} & \textbf{14.8} & \gooddelta{1.2} & \textbf{10.8} & \gooddelta{2.8} & \textbf{6.2} & \gooddelta{2.2} & \textbf{10.5} & \gooddelta{0.8} & \textbf{6.5} & \gooddelta{0.2} &\\[-1pt]

\bottomrule
\end{tabular}
}

\caption{
\textbf{X-ray OvOD performance under the Cross-Modality Transfer Evaluation (CMTE) setting} on DET-COMPASS (ours), PIXray~\cite{pixray}, PIDray~\cite{wang2021towards}, CLCXray~\cite{clcxray}, DvXray~\cite{dvxray}, and HiXray~\cite{HiXray} datasets.
We integrate \method into LaMI-DETR~\cite{du2024lami} using different gallery $\mathcal{G}$ compositions, from using only $\mathcal{D}^{\text{in-house}}_{\text{XRAY}}$ data (100/0) to exclusively $\mathcal{D}^{web}_{\text{RGB}}$ samples (0/100).
\method consistently improves the performance of LaMI-DETR.
}
\label{tab:main_exps_lami}

\end{table}
\begin{table}[t]
\setlength\extrarowheight{-2.5pt}
\centering
\resizebox{0.945\columnwidth}{!}{%
\begin{tabular}{@{}ll@{\hskip -8pt}|c@{\hskip 4pt}c@{\hskip 4pt}c@{\hskip 2pt}|c@{\hskip 4pt}c@{\hskip 4pt}c@{\hskip 4pt}@{}}
\toprule

& & %
\multirow{2}{*}{
    \makecell{
    {\smaller \textbf{Segment.\hspace{-5pt}}}
    }%
  }%
& %
\multirow{2}{*}{
    \makecell{
    {\smaller \textbf{LLM\hspace{-5pt}}}
    }%
  }%
& %
\multirow{2}{*}{
    \makecell{
    {\smaller \textbf{Features\hspace{-5pt}}}
    }%
  }%
& \multicolumn{3}{c}{\textbf{PIXray {\smaller (\textit{50/50})}}} \\[-1.5pt]
&  &  &  &  & \textbf{AP} & \textbf{AP50} & \textbf{AP75} \\[-4.5pt]

\midrule
\midrule

\multicolumn{2}{l|}{G-DINO~\cite{liu2025grounding}} & & & & 12.9 & 14.9 & 13.4\\[-4pt]

\midrule

\multirow{6}{*}{
    \plusours
  }%
& & SAM 2 & GPT-4 & DINOv2 & \textbf{25.4}\gooddelta{12.5} & \textbf{31.2}\gooddelta{16.3} & \textbf{26.7}\gooddelta{13.3}\\

& & -- & GPT-4 & DINOv2 & \textbf{22.0}\gooddelta{9.1} & \textbf{27.3}\gooddelta{12.4} & \textbf{22.7}\gooddelta{9.3}\\

& & SAM 2 & -- & DINOv2 & \textbf{20.8}\gooddelta{7.9} & \textbf{24.1}\gooddelta{9.2} & \textbf{21.4}\gooddelta{8.0}\\ 

& & SAM 2 & GPT-4 & DINO & \textbf{22.2}\gooddelta{9.3} & \textbf{27.6}\gooddelta{12.7} & \textbf{22.9}\gooddelta{9.5}\\ 

& & SAM 2 & LLaMA-3 & DINOv2 & \textbf{24.7}\gooddelta{11.8} & \textbf{30.1}\gooddelta{15.2} & \textbf{26.1}\gooddelta{12.7}\\ 

& & SAM & GPT-4 & DINOv2 & \textbf{25.1}\gooddelta{12.2} & \textbf{31.0}\gooddelta{16.1} & \textbf{26.4}\gooddelta{13.0}\\[-2pt]

\bottomrule
\end{tabular}
}

\caption{
\textbf{Ablation study of RAXO components on the PIXray~\cite{pixray} dataset (50/50 setting).} We integrate RAXO into G-DINO and analyze the impact of segmentation models, language models, and visual features. Results show that each component incrementally boosts performance, with the full RAXO configuration yielding the best results.
}

\label{tab:comp_ablation}

\end{table}

\section{Per-class AP}
\label{sec:sup_perclasaap}
\Cref{tab:category_ap} shows per-class AP on the PIXray dataset for G-DINO. \method consistently improves performance, especially on challenging categories with low baseline scores such as \textit{Pressure Vessel} (\gooddeltatext{52.3}), and \textit{Hammer} (\gooddeltatext{54.8}). In \cref{tab:per_cat_ap_compass}, we extend the per-category analysis to the DET-COMPASS dataset, analyzing the top-5 classes with the highest and lowest performance gains.
\method excels on items with distinctive shapes or strong cross-modal color shifts, while struggling with generic-shaped objects that provide limited cues under X-ray.

\section{Qualitative Analysis of the Material Transfer Mechanism}
\label{sec:sup_qual_mt}
\begin{figure*}[t]
\centering
\includegraphics[width=0.95\linewidth]{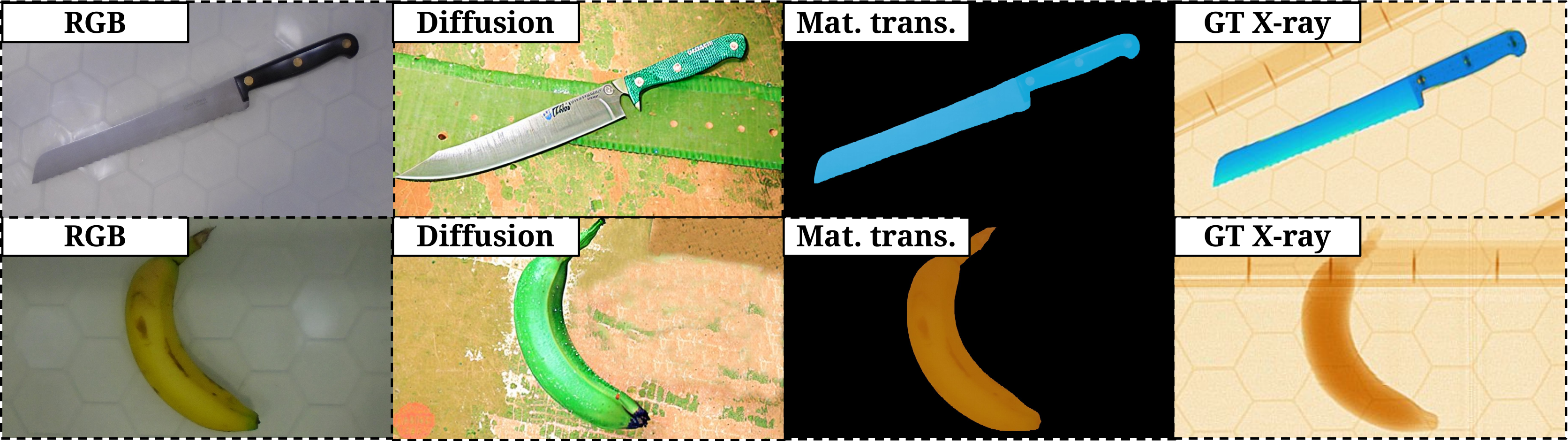}
\caption{\textbf{Qualitative comparison} between our material-transfer mechanism and a diffusion-based method~\cite{styleshot}.}
\label{fig:qualitative_material_transfer}
\end{figure*}

The core challenge that \method faces is tackling the domain gap between RGB and X-ray images without training or fine-tuning. The specific component we develop for this purpose is our material-transfer mechanism, whose results compared to a diffusion-based method~\cite{styleshot} can be found in \cref{fig:qualitative_material_transfer}.

\section{Qualitative Analysis of \method}
\label{sec:qualitative_method}

\begin{figure*}[t]
\centering
\includegraphics[width=0.89\linewidth]{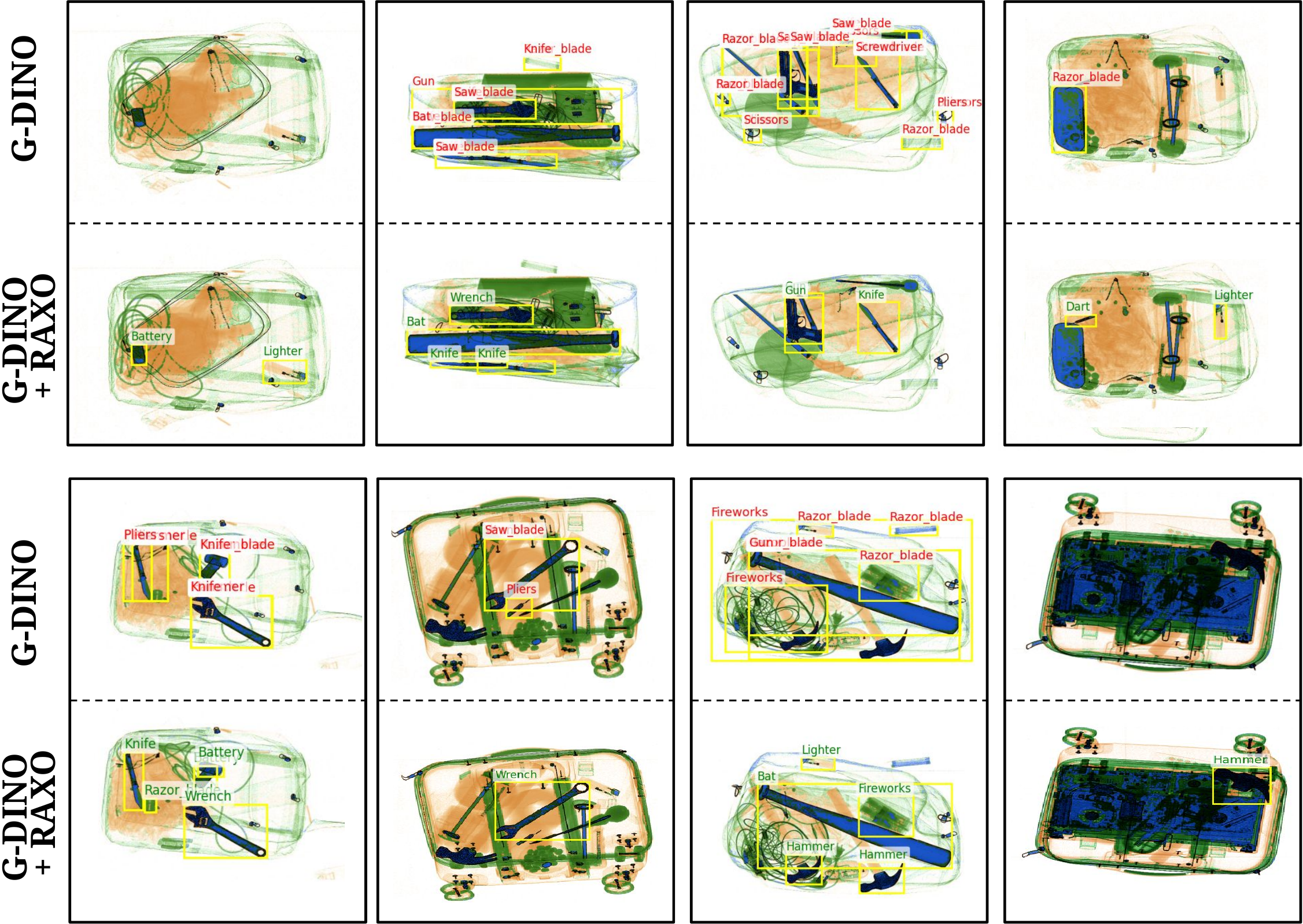}
\caption{\textbf{Qualitative comparison} of G-DINO~\cite{liu2025grounding} and G-DINO+\method.
}
\label{fig:supp-quali-comparison}
\end{figure*}

\cref{fig:supp-quali-comparison} presents qualitative visualizations of detected X-ray objects before and after applying \method with GroundingDINO~\cite{liu2025grounding} on the PIXray~\cite{pixray} dataset. For proper visualization, we display detections with a confidence score higher than 0.15 in both cases. These images lead to two key conclusions: (1) RAXO significantly improves the classification of detected proposals. In the baseline images, many objects are correctly localized but misclassified. RAXO successfully corrects these misclassifications by constructing robust visual descriptors. (2) The use of both background descriptors and the Descriptor Consistency Criterion (DCC) effectively eliminates false positives that do not correspond to actual X-ray objects. These observations strongly support the reliability of \method.

\newpage

\begin{table*}[!t]
\centering
\resizebox{0.95\linewidth}{!}{%
\begin{tabular}{lllllll}
\toprule
\multicolumn{7}{c}{\textbf{DET-COMPASS Categories}} \\
\toprule

abacus & abaya & amplifier & analog watch & apron & baby monitor & backpack \\
bag of sweets & baking dish & ballpoint & banana & Band Aid & baseball bat & baseball cap \\
bath towel & bathing cap & beanie & beer bottle & beer glass & bell pepper & belt \\
bib & bicycle helmet & bikini & binder & binoculars & bird feeder & biscuits \\
blowtorch & boardgame & book & book jacket & boot & bow tie & bowl \\
bowler hat & box cutter & bracelet & brassiere & bread knife & brush & bumbag \\
butternut squash & cable & caliper & camcorder & camera & can opener & candle \\
canned food & capo & cardigan & cards & carving knife & cassette & cassette player \\
cd drive & cellular telephone & cereal & chain & charger & chewing gum & chisel \\
chocolate & chocolate sauce & Christmas stocking & cigarettes & clarinet & coat hanger & cocktail shaker \\
coffee mug & coffeepot & colander & comb & combination lock & comic book & compact disc \\
condoms & corkscrew & cotton buds & cotton wool & cowboy hat & craft knife & crayon \\
crisps & crossword puzzle & crowbar & cucumber & dagger & denture & deodorant \\
diaper & digital watch & dinner jacket & dishrag & dressing gown & dvd player & e cigarette \\
e liquid & electric fan & electric toothbrush & empty & envelope & espresso maker & extension cord \\
face powder & fascinator & feather boa & first aid kit & floss & flute & fork \\
French loaf & frisbee & frying pan & fur coat & gaffer tape & game console & gameboy \\
gas canister & glove & glue gun & goggles & hacksaw & hair clippers & hair gel \\
hair spray & hairbrush & hammer & hand blower & handkerchief & hard disc & harmonica \\
hatchet & headphones & hearing aid & high heel & hook & hourglass & ipad \\
iPod & iron & jean & jersey & jewellery box & jigsaw puzzle & joystick \\
jumper & kettle & keys & kimono & kindle & kiwi & knee pad \\
knife & lab coat & ladle & lampshade & laptop & laser pointer & leather jacket \\
lemon & lens & lens cap & letter opener & lighter & lime & lipstick \\
lotion & loudspeaker & loupe & magazine & magnetic compass & maillot & mallet \\
marker & mask & matchstick & measuring cup & microphone & milk can & milk carton \\
mitten & mixing bowl & modem & mortar & mosquito net & mouse & mousetrap \\
mouthwash & multimeter & music stand & nail & nail clippers & nail file & nail scissors \\
necklace & notebook & orange & oxygen mask & padlock & paint can & paintbrush \\
pajama & paper towel & passport & pasta & pencil & pencil box & pencil sharpener \\
penknife & pepper grinder & perfume & pick & pickaxe & piggy bank & pill bottle \\
pillow & plane & plastic bag & plate & plate rack & pliers & plunger \\
Polaroid camera & polo shirt & pomegranate & poncho & pop bottle & pot & power drill \\
power socket & power supply & prayer rug & quill & quilt & quilted jacket & radio \\
rasp & razor & razor blades & recorder & red wine & reflex camera & remote control \\
rice & roll of sweets & roller skate & rubber eraser & rubber gloves & rubik cube & rugby ball \\
rugby shirt & rule & running shoe & safety pin & salad bowl & saltshaker & sandal \\
sandwich & sarong & saucepan & saw & sax & scale & scarf \\
scissors & screw & screwdriver & secateurs & sellotape & sewing machine & shampoo \\
shaver & shawl & shirt & shorts & shovel & shower cap & sieve \\
ski mask & skipping rope & sleeping bag & slide & slotted spoon & smartphone & snorkel \\
soap & soap dispenser & sock & solder & soldering iron & sombrero & soup bowl \\
spatula & spectacles & spirit level & splitter block & spoon & spotlight & staple gun \\
stapler & stethoscope & stockings & stole & stopwatch & strainer & strings \\
stylophone & suit & sunglasses & sunscreen & swab & sweatshirt & swimming trunks \\
switch & syringe & table lamp & tampon & tape measure & tea towel & teapot \\
teaspoon & teddy & telephone & telescope & tennis ball & thermals & thermometer \\
thermos & tin of sweets & toaster & toilet tissue & toner cartridge & toothbrush & toothpaste \\
top hat & torch & tracksuit & tray & tripod & tuner & ukulele \\
umbrella & underpants & vacuum & vase & velvet & vinyl record & violin \\
waffle iron & walking boot & wall clock & wallet & washbag & water bottle & water jug \\
wellington boot & wet wipes & whetstone & whistle & wig & wineglass & wire wool \\
wirecutter & wok & wooden spoon & wool & wrench & wrist guard \\

\bottomrule
\end{tabular}
}
\caption{
Category names of DET-COMPASS.
}
\label{tab:categories_compass}
\end{table*}
\newpage

\begin{table*}[!t]
\begin{tcolorbox}[breakable, enhanced jigsaw,title={(1): Material-database clustering prompt}]
``You are a computer expert specializing in material classification. Your task is to analyze a given list of objects, determine their primary material composition, and group them accordingly. \\

Instructions: \\
Identify the main materials present among the objects (e.g., metal, organic, inorganic, plastic, ceramic, etc.). Assign each object to the most appropriate material category. Each object should belong to only one category based on its primary composition. Return the results in JSON format, where the keys are material categories, and the values are lists of objects belonging to those categories.\\

Example: \\
Input: Objects: gun, bat, pressure vessel, beer glass, fur coat, lemon

Expected Output (JSON):

  metal: [gun, bat],\\
  inorganic: [pressure vessel, beer glass],\\
  organic: [fur coat, lemon]\\

Now, classify the following list of objects: \{$D^{\text{in-house}}$\}. Return only the json format.''
\end{tcolorbox}

\begin{tcolorbox}[breakable, enhanced jigsaw,title={(2): Object material identification prompt}]
``You are a computer vision assistant. Given a \{\textit{object}\}, classify it into one of the following materials: \{$\mathcal{M}.$\textit{materials\_names}\}. Return only the material. You must always select one.''
\end{tcolorbox}

\caption{\textbf{Prompts used for material clustering and retrieval}. (1) The clustering prompt provided to GPT-4 to group $\mathcal{C}^{\text{in-house}}$ into material categories. (2) The retrieval prompt used to query $\mathcal{M}$ and infer the expected material of unknown RGB objects.}
\label{tab:promts}
\end{table*}
\newpage

\end{document}